\documentclass[letterpaper,11pt]{article}

\usepackage[utf8]{inputenc}
\usepackage[T1]{fontenc}
\usepackage{times}
\usepackage[margin=1in,top=0.45in]{geometry}
\usepackage{microtype}

\usepackage{amsmath,amssymb,amsfonts,amsthm}
\usepackage{mathrsfs}
\usepackage{dsfont}

\usepackage{amsmath,amsfonts,bm}

\def\eqref#1{equation~\ref{#1}}

\def\1{\bm{1}}

\def\rvv{{\mathbf{v}}}

\def\ve{{\bm{e}}}

\def\vg{{\bm{g}}}
\def\vh{{\bm{h}}}

\def\vu{{\bm{u}}}

\def\vz{{\bm{z}}}

\DeclareMathAlphabet{\mathsfit}{\encodingdefault}{\sfdefault}{m}{sl}
\SetMathAlphabet{\mathsfit}{bold}{\encodingdefault}{\sfdefault}{bx}{n}

\newcommand{\E}{\mathbb{E}}
\newcommand{\Ls}{\mathcal{L}}

\usepackage{graphicx}
\usepackage[export]{adjustbox}
\usepackage{tabularx}
\usepackage{booktabs}
\usepackage{multirow}
\usepackage{makecell}
\usepackage{longtable}
\usepackage{wrapfig}
\usepackage{float}
\usepackage{array}
\usepackage{subcaption}

\usepackage{enumitem}
\usepackage{xspace}
\usepackage[table]{xcolor}
\usepackage[most]{tcolorbox}
\usepackage{etoc}
\usepackage{titlesec}
\usepackage{todonotes}

\usepackage{natbib}
\usepackage{url}
\usepackage[colorlinks=true,linkcolor=cyan,citecolor=cyan,filecolor=magenta,urlcolor=blue]{hyperref}
\usepackage{cleveref}

\newcommand{\method}{FlexUMM\xspace}

\newcommand{\cmark}{$\checkmark$}
\newcommand{\xmark}{$\times$}

\newcommand{\impr}[1]{\,\textcolor{cyan}{\scriptsize$\uparrow$#1\%}}

\setlist{leftmargin=5mm}
\definecolor{abstractbg}{RGB}{230,242,250}
\definecolor{abstractborder}{RGB}{200,210,220}

\titleformat{\section}
  {\sffamily\bfseries\Large}
  {\thesection}
  {1em}
  {}

\renewenvironment{abstract}
{\begin{center}
\begin{tcolorbox}[
    colback=abstractbg,
    colframe=abstractborder,
    boxrule=0.6pt,
    arc=6pt,
    width=\textwidth,
    left=8pt,
    right=8pt,
    top=8pt,
    bottom=8pt]
}
{\end{tcolorbox}
\end{center}}

\newcommand{\papertitle}{%
  \sffamily\bfseries\fontsize{16}{20}\selectfont
  \method: Flexible Representation Learning for Unified Multimodal Models
}
\newcommand{\paperauthors}{%
  \sffamily
  Yinyi Luo$^{1}$\footnotemark[1], Haoyue Bai$^{2}$, Yi Ge$^{3}$,\\
  Nilton Cesar Rojas Vales$^{2}$, Marios Savvides$^{1}$, Jindong Wang$^{2}$\footnotemark[2]
}
\newcommand{\paperaffiliations}{%
  \sffamily
  $^{1}$Carnegie Mellon University \quad
  $^{2}$William \& Mary \quad
  $^{3}$University of Maryland
}
\begin{document}
\thispagestyle{empty}

\noindent\includegraphics[height=.6cm]{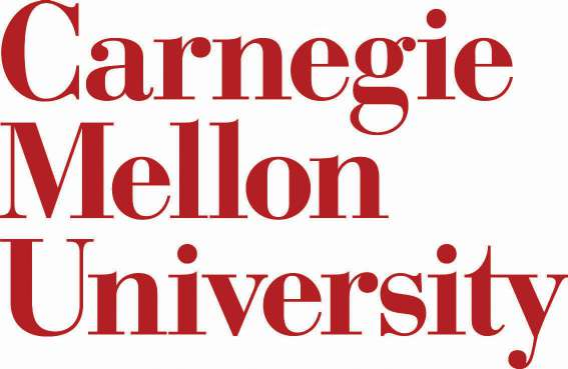}~%
\includegraphics[height=.6cm]{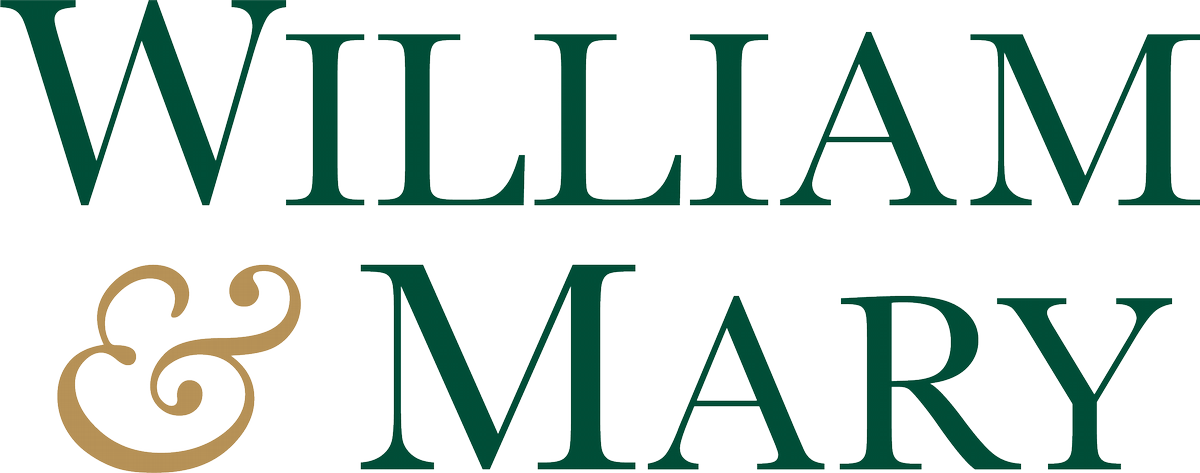}~%
\includegraphics[height=.6cm]{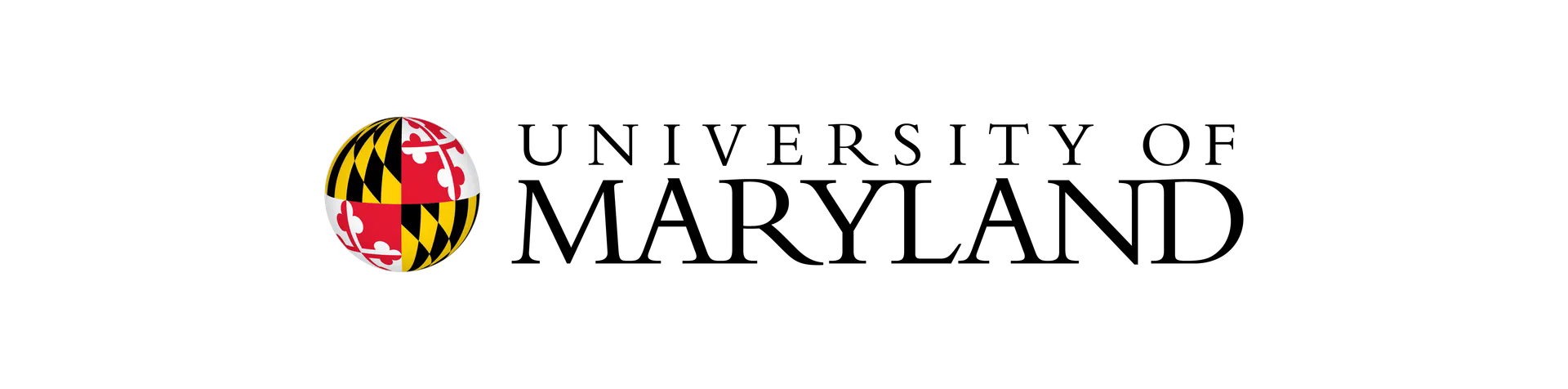}\par
\noindent\rule{\textwidth}{0.8pt}
\renewcommand{\thefootnote}{\fnsymbol{footnote}}
\begin{center}
  {\papertitle\par}
  {\large \paperauthors\par}
  {\normalsize \paperaffiliations\par}
\end{center}

\footnotetext[1]{Contact: yinyil@andrew.cmu.edu}
\footnotetext[2]{Corresponding authors: jdw@wm.edu}

\begin{abstract}
Unified multimodal models (UMMs) can integrate understanding and generation within a single backbone.
However, it is commonly known that these two capabilities can cause conflicting representations, leading to a central question: how representation and computation should be shared across functionalities?
Existing UMMs often rely on a largely fixed structure, where different inputs and functionalities follow the same depth and parameter-sharing pattern.
We argue that this assumption is too restrictive: more effective unification requires sharing the computation that supports cross-functional communication while specializing the capacity that becomes task- or input-specific.
We introduce \method, a flexible UMM framework that dynamically decides when to preserve shared computation, when to route specialized capacity, and when to stop further processing.
Evidence shows that this flexible allocation improves UMMs along two complementary axes.
It improves accuracy by reducing harmful competition between functionalities, and it improves efficiency by avoiding unnecessary computation once an input representation is sufficiently mature.
Beyond simple architectural unification from the pre-trained models, \method works as a model-agnostic, \textit{functional} plug-in to make existing UMMs more unified, stronger, and efficient. Code is available at: \url{https://github.com/AIFrontierLab/TorchUMM/tree/main/src/umm/post_training/flexumm}.
\end{abstract}

\addtocontents{toc}{\protect\setcounter{tocdepth}{-1}}

\section{Introduction}

Unified multimodal models (UMMs) are built around an appealing idea: understanding and generation can use one backbone instead of being served by two separate models \citep{wang2022ofa,lu2022unified,sun2024generative,deng2025emerging}.
Using the same shared architecture, one UMM can do both generation and understanding tasks with the same network.
This unification, however, creates a fundamental challenge in knowledge representation and computation allocation \citep{wu2026uncovering,kil2026efficient}.
A shared backbone must simultaneously support different capabilities, yet not every input requires the same computation, and not every component of the representation should be shared across capabilities \citep{han2026turning,kang2026transferability}.

\suppressfloats[t]
\begin{figure*}[b!]
\centering
\captionsetup[subfigure]{font=scriptsize,skip=.5pt}
\noindent
\begin{minipage}[b]{.60\linewidth}
  \centering
  \begin{subfigure}[b]{\linewidth}
    \centering
    \includegraphics[width=\linewidth]{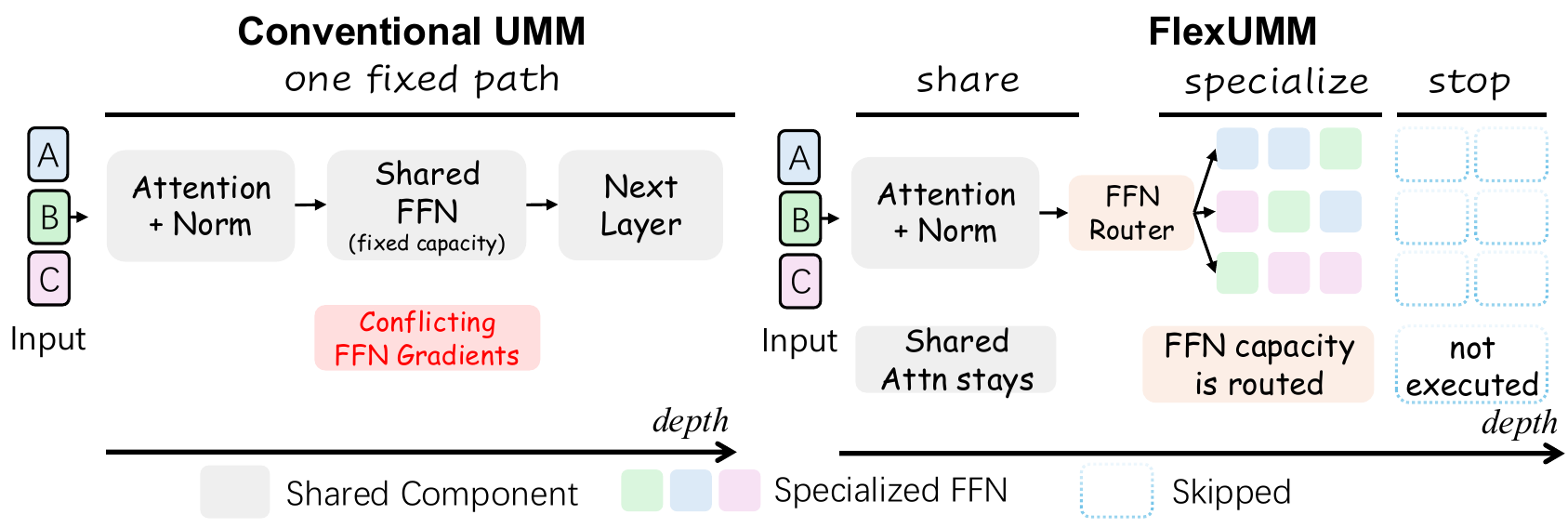}
    \caption{Shared communication versus routed FFN capacity.}
    \label{fig:teaser-a}
  \end{subfigure}
\end{minipage}\hfill%
\begin{minipage}[b]{.28\linewidth}
  \centering
  \begin{subfigure}[b]{\linewidth}
    \centering
    \includegraphics[width=.94\linewidth]{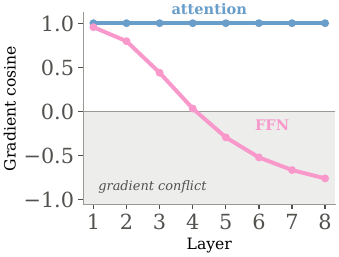}
    \caption{Cross-functional gradient cosine.}
    \label{fig:teaser-b}
  \end{subfigure}
\end{minipage}
\caption{\method adapts both who computes and how much computation an input receives.}
\label{fig:teaser}
\end{figure*}

Existing UMMs largely rely on a fixed computational structure (\Cref{fig:teaser-a}), leaving both the allocation of capacity across capabilities and depth across inputs predetermined \citep{kil2026efficient,qiu2026new}.
Such a rigid allocation can lead to two forms of mismatch: computation may be spent on inputs or capabilities that do not require it, while shared components may be overused even when their computation is not equally beneficial to different capabilities \citep{he2025understanding, wu2026uncovering, ye2026understanding}.
This limitation raises a central question: \textit{how should a UMM adapt its representation and computation to both capabilities more adaptively?} Addressing this question requires resolving two closely related challenges.

\begin{wrapfigure}{r}{.48\textwidth}
\centering
\begin{subfigure}[t]{.48\linewidth}
  \centering
  \includegraphics[width=.94\linewidth]{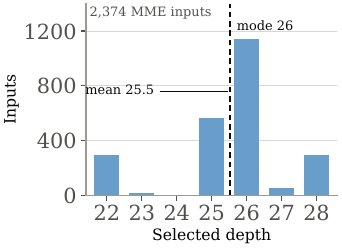}\\[-.02in]
  \caption{Selected depths.}
  \label{fig:depth-efficiency-a}
\end{subfigure}\hfill%
\begin{subfigure}[t]{.48\linewidth}
  \centering
  \includegraphics[width=.94\linewidth]{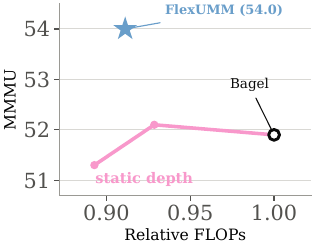}\\[-.02in]
  \caption{Quality--efficiency behavior.}
  \label{fig:depth-efficiency-b}
\end{subfigure}
\caption{Input-adaptive depth and quality--efficiency behavior.}
\label{fig:depth-efficiency}
\end{wrapfigure}

\textit{First, computation depth should adapt to the input and capability.}
Not every input requires the full depth of a deep network \citep{graves2016adaptive,elbayad2019depth}. In a UMM, however, the required depth depends not only on the input but also on the capability being executed \citep{kang2026transferability,su2026generation}.
The depth at which a representation becomes sufficient for understanding should not coincide with the depth required for image generation, even though both capabilities operate on the same backbone.
A fixed execution depth therefore imposes the same computational budget on inputs with different requirements.
It is evident even within a single functionality: using a learned policy on Bagel, all MME \citep{fu2026mme} inputs are distributed across six distinct depths (\Cref{fig:depth-efficiency-a}).
Moreover, the quality--efficiency behavior in \Cref{fig:depth-efficiency-b} suggests that computation beyond the sufficient depth can introduce unnecessary interference and reduce accuracy, rather than simply incurring redundant FLOPs.
These observations motivate input-adaptive depth allocation that determines whether further computation remains useful.

\textit{Second, computational capacity must adapt to the capability.}
While depth determines how far computation proceeds, it does not determine which components should be shared or specialized.
Understanding and generation rely on a common backbone, but their objectives can impose conflicting demands on the same parameters \citep{han2026turning,wu2026uncovering}.
Separating the capabilities entirely would reduce such interference, but would also discard the cross-capability communication that motivates unified modeling \citep{kang2026transferability,su2026generation}.
The challenge is therefore to preserve shared computation where the capabilities benefit from interaction while allowing specialization where their objectives diverge \citep{peng2026unified,qiu2026new}.
Our controlled diagnostic, under the task construction described in the appendix, suggests a structured sharing pattern: attention gradients remain aligned across functionalities throughout the network, whereas FFN gradients increasingly diverge with depth (\Cref{fig:teaser-b}, Appendix \ref{sec:sharing-analysis}). This is a design hypothesis rather than a direct measurement of every released backbone, and it suggests that communication and transformation capacity should not be treated uniformly.
Communication benefits from sharing while transformation capacity requires adaptive specialization.
These interactions also have implications beyond accuracy and efficiency.
If shared computation is adapted without accounting for capability-specific demands, improvements in one functionality may come with safety regressions in another.

We introduce \textbf{\method}, a flexible computation allocation method that adapts both capacity and depth on a shared UMM backbone.
\method uses capability-adaptive allocation to preserve shared attention and selectively specialize FFN capacity through per-token routing, and depth-adaptive allocation to determine whether further computation is needed through per-sequence early exit.
Together, these mechanisms form a progressive \textit{share, specialize, and stop} policy that dynamically allocates computation across capabilities and inputs (\Cref{fig:teaser-a}).
\method is implemented as a lightweight adaptation on top of the pretrained checkpoint, adding only routers, halting heads and low-rank residual banks, with only 0.381\% increase in model size.

Our contributions are summarized as follows:
\begin{itemize}[leftmargin=2em]
\setlength\itemsep{0em}
    \item We identify two complementary sources of conflicts in UMMs: input-dependent depth and capability-dependent capacity. Our analysis shows that attention provides a shared communication pathway, while FFN capacity exhibits increasing functional divergence across layers.
    \item We propose \method, which jointly enables token-level capacity allocation and sequence-level depth adaptation within a unified backbone, allowing computation to be shared, specialized, or terminated according to the input and capability.
    \item \method improves both multimodal understanding and generation while providing an adjustable accuracy-compute trade-off at inference time, with negligible extra computation.
\end{itemize}

\section{Related Work}

\textbf{Unified multimodal models.}
UMMs combine multimodal understanding and generation within a single architecture, with different designs for shared representations, tokenizers, and computation paths \citep{xie2025show,chen2025janus,deng2025emerging, peng2026unified}. Recent studies further examine how architectural choices affect the interaction between functionalities \citep{kang2026transferability, qiu2026new, wu2026uncovering, luo2026latentumm,luo2026self,bai2026unipath}. These models validate unification, but their computation paths remain largely fixed, with sharing patterns and depth predetermined before inference.

\textbf{Adaptive sharing and specialization.}
Multi-task models mitigate task interference by mixing task-specific representations, routing through shared and private experts, or learning layerwise sharing policies \citep{misra2016cross, ma2018modeling, tang2020progressive, sun2020adashare, rahimian2023dynashare}. UMMs have also introduced function-specific computation to reduce conflicts: Janus decouples visual encoding for understanding and generation, while more recent work specializes conflicting visual computation while preserving semantic interaction \citep{wu2025janus, wu2026uncovering}. These approaches mainly target conventional multi-task learning or rely on predefined specialization, rather than dynamically allocating shared and specialized computation.

\textbf{Adaptive computation.}
Adaptive computation methods reduce visual processing through token selection, pruning, or merging \citep{ryoo2021tokenlearner, rao2021dynamicvit, bolya2022token}. PruMerge, FastV, SparseVLM, DyVTE, and HiDrop \citep{shang2025llava,chen2024image,zhang2024sparsevlm, wu2026accelerating, wu2026hidrop} exploit visual-token redundancy or vary token participation across layers, while MuE, Mixture-of-Depths, and MoR adapt executed depth or recursive computation \citep{tang2023you, raposo2024mixture, bae2026mixture}. These methods adapt token participation or executed depth; \method additionally adapts whether computation is shared or specialized.

\section{Method}

\begin{figure*}[t!]
  \centering
  \includegraphics[width=.9\linewidth]{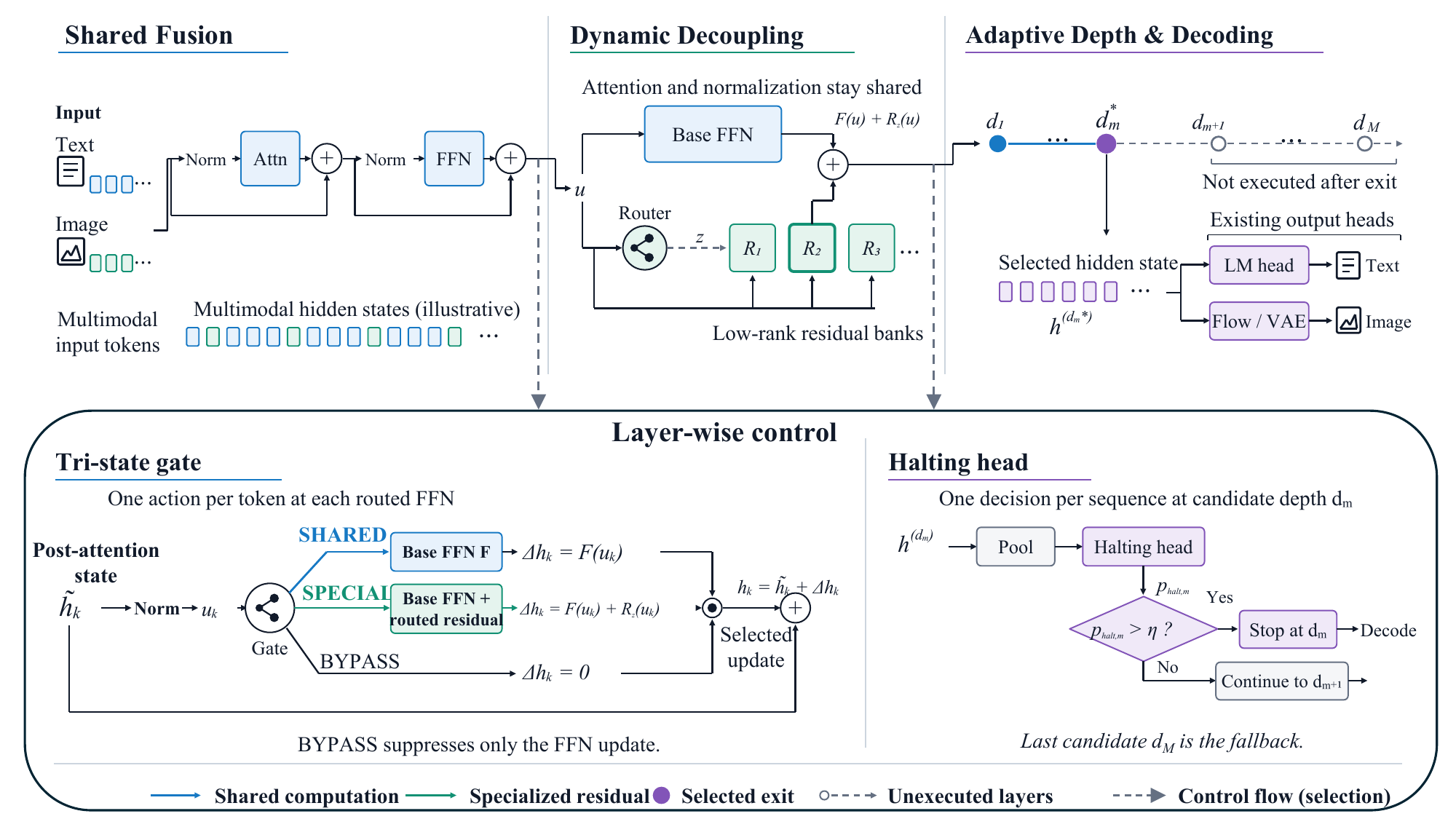}
  
  \caption{\textbf{\method separates three responsibilities in the implementation.}
  The shared decoder stack supplies attention and normalization for every route.
  A tri-state capacity gate selects \textsc{Share}, \textsc{Decouple}, or
  \textsc{Exit} for each token at each routed FFN.  A separate candidate-depth
  router decides, at the sequence level, whether the remaining decoder layers
  are still needed.}
  \label{fig:pipeline}
  
\end{figure*}

As illustrated in \Cref{fig:pipeline}, \method introduces an input-adaptive computation method for UMMs. The model preserves a shared backbone for multimodal communication while adapting both the amount of capacity allocated to individual tokens and the computation depth required by each input.
It consists of three stages:
1) Shared Fusion: All tokens undergo the same attention, normalization, and FFN (feed-forward network) to establish a common multimodal representation.
2) Adaptive Specialization: The shared backbone remains intact, while a Tri-state Gate determines whether each token uses the shared FFN alone, receives an additional specialized residual transformation, or bypasses the FFN.
3) Adaptive Depth: A sequence-level Halting Head determines whether the current representation is sufficient, allowing different inputs to terminate computation at different depths.
Thus, \method adapts computation along two complementary dimensions: capacity at the token level and depth at the sequence level, while retaining a shared multimodal communication pathway.

\subsection{Shared Communication and Adaptive Capacity}
\label{sec:capacity}
Let $\vh^{(\ell-1)} \in \mathbb{R}^{n\times d}$ denote the hidden states entering layer $\ell$, where $n$ is the sequence length and $d$ is the hidden dimension. Each layer follows the standard Transformer backbone with multi-head attention and normalization shared across modalities: 
\begin{equation}
\tilde{\vh}^{(\ell)} = \vh^{(\ell-1)} + \mathrm{Attn}^{(\ell)} \left( \mathrm{Norm}_1^{(\ell)} \left( \vh^{(\ell-1)} \right) \right), \qquad \vu^{(\ell)} = \mathrm{Norm}_2^{(\ell)} \left( \tilde{\vh}^{(\ell)} \right),
\end{equation}
where $\vu_k^{(\ell)}\in\mathbb{R}^{d}$ and $\tilde{\vh}_k^{(\ell)}\in\mathbb{R}^{d}$ represent the normalized and residual state for token $k$, respectively.

\textbf{Adaptive capacity allocation.}
\label{sec:tri-state}
At a given layer $\ell$, a Tri-state Gate dynamically selects an
action $a_k^{(\ell)} \in \mathcal{A}
= \{ \mathrm{SHARED}, \mathrm{SPECIALIZED}, \mathrm{BYPASS} \}$
to allocate FFN capacity for token $k$. Specifically,
$\mathrm{SHARED}$ routes the token through the shared base FFN,
$\mathrm{SPECIALIZED}$ augments the shared FFN with a learned
decoupled residual subset, and $\mathrm{BYPASS}$ skips the current
FFN update. The gate predicts the action based on the token
representation and routing context, enabling input-adaptive
capacity allocation without requiring manually annotated token
attributes.

Let $s_{k,a}^{(\ell)}$ denote the gate logit for action $a$.
The action probability and inference decision are:
\begin{equation}
\pi_{k,a}^{(\ell)}
=
\frac{
\exp\left(s_{k,a}^{(\ell)}/\tau\right)
}{
\sum_{a'\in\mathcal{A}}
\exp\left(s_{k,a'}^{(\ell)}/\tau\right)
},
\qquad
a_k^{(\ell)}
=
\arg\max_{a\in\mathcal{A}}
s_{k,a}^{(\ell)}.
\label{eq:gate_prob}
\end{equation}

For the $\mathrm{SPECIALIZED}$ action, a learned routing head
selects a residual-bank index $z_k^{(\ell)}$ from the token
representation and routing context. The index identifies a
learned residual subset rather than a manually assigned token
label. The overall output state is
$\vh_k = \tilde{\vh}_k + \Delta\vh_k$, where the token-level FFN update $\Delta \vh_k$ is defined as
\begin{equation}
\Delta \vh_k =
\begin{cases}
F\left(\vu_k\right),
& a_k^{(\ell)}=\mathrm{SHARED}, \\[2pt]
F\left(\vu_k\right)
+
R_{z_k^{(\ell)}}\left(\vu_k\right),
& a_k^{(\ell)}=\mathrm{SPECIALIZED}, \\[2pt]
\mathbf{0},
& a_k^{(\ell)}=\mathrm{BYPASS}.
\end{cases}
\label{eq:capacity_update}
\end{equation}
Here, $F$ denotes the shared base FFN, while
$R_{z_k^{(\ell)}}$ denotes the selected learned residual subset.
The residual subsets are optimized jointly with the routing
mechanism during post-training, providing additional
input-adaptive capacity while preserving the shared FFN as the
common foundation.

\subsection{Adaptive Computation Depth}
\label{sec:depth}

While the capacity gate dynamically allocates computation within each
layer, the effective computation depth may vary across inputs. We
therefore introduce a depth router that selects from a predefined set
of candidate exit depths
$\mathcal{D}=\{d_1,\ldots,d_M\}$, where
$d_1<\cdots<d_M$. Here, $m$ corresponds to a configured
candidate depth rather than a continuously learned computation depth.

At candidate depth $d_m$, the depth router receives the current shared
representation and produces a halting logit $r_m$. The corresponding
halting probability is defined as $p_{\mathrm{halt},m}
=
\sigma\left(\frac{r_m}{\tau_{\mathrm{depth}}}\right),$
where $\tau_{\mathrm{depth}}$ denotes the temperature of the depth
router and $\sigma(\cdot)$ is the sigmoid function. We distinguish
$\tau_{\mathrm{depth}}$ from the temperature used by the capacity gate,
as the two temperatures govern different routing decisions. In our
implementation, $\tau_{\mathrm{depth}}=1.0$.

\textbf{Sequential Exit Distribution.}
To jointly optimize the candidate exits, we define a sequential exit
distribution based on the halting probabilities:
\begin{equation}
q_m =
\begin{cases}
p_{\mathrm{halt},m}
\displaystyle\prod_{r=1}^{m-1}
(1-p_{\mathrm{halt},r}), & m<M,\\[6pt]
\displaystyle\prod_{r=1}^{M-1}
(1-p_{\mathrm{halt},r}), & m=M.
\end{cases}
\label{eq:exit_prob}
\end{equation}
The final candidate serves as a mandatory fallback, ensuring that
$\sum_{m=1}^{M}q_m=1$. The resulting expected computation depth is $\mathbb{E}[\mathcal{D}]
=
\sum_{m=1}^{M}q_m d_m.$
During training, the full-depth candidate provides the reference
computation for assessing the quality of earlier exits. Each candidate
is evaluated according to its task objective, and candidates satisfying
a predefined quality constraint relative to the full-depth reference
are considered eligible for early termination. The depth router is
optimized to favor earlier eligible candidates while retaining the
full-depth candidate as a fallback. This training procedure encourages
input-adaptive computation allocation without requiring the model to
learn an unrestricted continuous exit depth.

\textbf{Inference-Time Exit Policy.}
At inference, the model evaluates the configured candidate depths in
ascending order and terminates at the first candidate whose halting
probability exceeds a threshold:
\begin{equation}
m^\star
=
\min\left\{
m\geq m_{\min}
\;\middle|\;
p_{\mathrm{halt},m}\geq\eta_\beta
\right\},
\label{eq:inference_exit}
\end{equation}
where $m_{\min}$ specifies the earliest eligible exit and
$\eta_\beta$ is an inference-time threshold controlling the
computation-quality trade-off. If no earlier candidate satisfies the
criterion, the model falls back to the final candidate $m=M$.
In our experiments, we set the inference threshold to
$\eta_\beta=0.50$.

\subsection{Training and Inference}
\label{sec:objective}

\method jointly optimizes task performance and computational efficiency
through token-level capacity allocation and sequence-level depth
adaptation. The overall objective is defined as
\begin{equation}
\mathcal{L}
=
\mathcal{L}_{\mathrm{task}}
+
\lambda_{\mathrm{cap}}\mathcal{L}_{\mathrm{cap}}
+
\lambda_{\mathrm{depth}}\mathcal{L}_{\mathrm{depth}},
\label{eq:objective}
\end{equation}
where $\lambda_{\mathrm{cap}}$ and $\lambda_{\mathrm{depth}}$ are hyperparameters.

\textbf{Task Objective.}
To preserve the native task interfaces of the unified model, we reuse the original output heads at each candidate depth rather than introducing task-specific prediction heads. 
Specifically, the language-model head is used during the autoregressive decoding process for understanding, whereas the flow-matching head is used during the generation decoding process.
The task loss at candidate depth $d_m$ is defined as
\begin{equation}
\ell_m^{\mathrm{task}}
=
\begin{cases}
\mathcal{L}_{\mathrm{CE}}
\left(\mathrm{LMHead}(\vg_m),y\right),
& \text{understanding},\\[4pt]
\left\|
\mathrm{FlowHead}(\vg_m)-\rvv
\right\|_2^2,
& \text{generation},
\end{cases}
\label{eq:candidate_task_loss}
\end{equation}
where $\rvv$ denotes the target flow-matching velocity and
$\mathcal{L}_{\mathrm{CE}}$ denotes the cross-entropy loss for the
understanding task. The expected task loss across candidate exits is
then formulated as
\begin{equation}
\mathcal{L}_{\mathrm{task}}
=
\sum_{m=1}^{M}
q_m\ell_m^{\mathrm{task}},
\label{eq:task_loss}
\end{equation}
where $q_m$ is the sequential exit probability defined in
Eq.~(\ref{eq:exit_prob}). This formulation allows the depth router to
account for task quality across candidate computation depths while
retaining the native prediction interfaces of the underlying model.

\textbf{Capacity Regularization.}
The capacity regularizer encourages efficient allocation of token-level
computation across the adaptive layers. Let
$\mathcal{A}=\{\mathrm{SHARED},\mathrm{SPECIALIZED},\mathrm{BYPASS}\}$
denote the available actions, and let $c_a$ represent the normalized
relative computational cost associated with action $a$. We define
\begin{equation}
\mathcal{L}_{\mathrm{cap}}
=
\frac{1}{Ln}
\sum_{\ell=1}^{L}
\sum_{k=1}^{n}
\sum_{a\in\mathcal{A}}
c_a\,\pi_{k,a}^{(\ell)},
\label{eq:cap_loss}
\end{equation}
where $\pi_{k,a}^{(\ell)}$ denotes the action probability at layer
$\ell$ for token $k$, as defined in Eq.~(\ref{eq:gate_prob}).
The relative costs are determined according to the computation
associated with each action, with
$c_{\mathrm{BYPASS}} \leq c_{\mathrm{SHARED}}
\leq c_{\mathrm{SPECIALIZED}}$. The cost values are normalized with
respect to a common reference to ensure that the regularizer reflects
the relative computational burden of the available routes.

\paragraph{Depth Regularization.}
To promote input-adaptive early termination, we penalize the expected
computation depth normalized by the maximum candidate depth:
\begin{equation}
\mathcal{L}_{\mathrm{depth}}
=
\frac{\mathbb{E}[\mathcal{D}]}{d_M}
=
\frac{1}{d_M}
\sum_{m=1}^{M}q_m d_m.
\label{eq:depth_loss}
\end{equation}
This regularization encourages the router to favor earlier candidate
exits when they provide sufficient task quality, while retaining the
full-depth candidate as a fallback.

\paragraph{Inference and Computational Efficiency.}
During inference, \method evaluates candidate depths in ascending order
and terminates execution at the first eligible candidate according to
the halting policy in Eq.~(\ref{eq:inference_exit}). Unlike uniformly
deep execution, this adaptive policy allows samples to exit at
earlier depths, reducing unnecessary decoder computation and improving
inference efficiency.

By jointly adapting token-level capacity and sequence-level depth,
\method reduces redundant computation while preserving the shared
multimodal processing pathway. Token-level capacity allocation adjusts
the computational route within each layer, while sequence-level early
exiting reduces the number of executed layers, enabling more efficient
inference.

\section{Experiments}

We perform extensive experiments on effectiveness, efficiency, and explanations of \method.

\subsection{Experimental Setup}
\label{sec:setup}

\textbf{Backbone models.}
We evaluate \method on three UMM backbones with different sharing structures: Bagel \citep{deng2025emerging}, Janus-Pro \citep{chen2025janus}, and Harmon \citep{wu2025harmonizing}, with Bagel serving as the primary backbone. 

\textbf{Post-training data.}
\method is trained applied to each released checkpoint using ScienceQA \citep{lu2022learn}, AI2D \citep{kembhavi2016diagram}, TextVQA \citep{singh2019towards}, MathV-360K \citep{shi2024math}, and InstructPix2Pix \citep{brooks2023instructpix2pix}.

\textbf{Evaluation benchmarks and metrics.}
For understanding, we use MMMU \citep{yue2024mmmu}, MathVista \citep{lu2024mathvista}, MMStar \citep{chen2024we}, and MMBench \citep{liu2024mmbench}.
For generation, GenEval \citep{ghosh2023geneval} and WISE \citep{niu2025wise}.
For unified evaluation, Unison \citep{liu2026unison} and Unified-Bench \citep{yan2026unified}.

\textbf{Implementation details.}
Only the routers, halting heads, and low-rank residual banks are trained; the backbone configurations and trainable parameter counts are summarized in Appendix~\Cref{tab:hparams}.
All experiments are conducted using TorchUMM \citep{luo2026torchumm} for fair comparison.
More details in Appendix \ref{sec:configuration}.

\subsection{Main Results}
\label{sec:main-results}

\textbf{\method improves understanding accuracy over existing UMMs.}
\Cref{tab:main} compares \method with existing unified models across understanding and generation benchmarks. 
\method achieves the best results on all five understanding benchmarks and consistently improves both generation metrics over its Bagel backbone, while remaining behind generation-specialized models. 
These gains demonstrate the benefit of adaptive computation allocation without increasing model scale.

\begin{table}[h]
\centering
\caption{Main results. \method is applied on top of the Bagel backbone; all other rows are released unified models evaluated under the same protocol. Best results are in \textbf{bold}.}
\label{tab:main}
\small
\setlength{\tabcolsep}{4pt}
\resizebox{\linewidth}{!}{
\begin{tabular}{lcccccc|cc}
\toprule
\multirow{2}{*}{Model} & \multirow{2}{*}{Params} & \multicolumn{5}{c|}{Understanding} & \multicolumn{2}{c}{Generation} \\ \cline{3-7} \cline{8-9}
 &  & MMMU & MMBench-CN & MMBench-EN & MathVista & MMStar & GenEval & WISE \\ \midrule
Bagel \citep{deng2025emerging} & 14B & 51.9 & 80.94 & 82.65 & 71.3 & 63.20 & 78.81 & 0.4044 \\
Janus-Pro \citep{chen2025janus} & 7B & 40.9 & 69.93 & 64.84 & 43.8 & 47.93 & 79.02 & 0.3900 \\
Janus \citep{wu2025janus} & 1.3B & 27.3 & 37.19 & 27.69 & 26.6 & 26.67 & 40.04 & 0.2222 \\
Janus Flow \citep{ma2025janusflow} & 1.3B & 29.0 & 60.31 & 64.16 & 34.8 & 39.13 & 49.99 & 0.2964 \\
Show-O2 \citep{xie2026show} & 7B & 47.9 & 76.68 & 77.25 & 51.5 & 54.53 & 59.87 & 0.3595 \\
Show-O2 \citep{xie2026show} & 1.5B & 37.1 & 61.04 & 65.09 & 37.9 & 43.87 & 55.49 & 0.3349 \\
Show-O \citep{xie2025show} & 1.3B & 26.1 & 11.01 & 43.48 & 29.0 & 27.87 & 65.06 & 0.3037 \\
Emu3 \citep{wang2024emu3} & 8B & 31.4 & 45.66 & 59.53 & 44.9 & 43.60 & 45.76 & 0.3373 \\
Emu3.5 \citep{cui2025emu3} & 34B & 29.2 & 18.13 & 17.56 & 41.7 & 30.27 & 81.83 & \textbf{0.6331} \\
Omnigen2 \citep{wu2026omnigen2} & 7B & 46.0 & 75.43 & 76.36 & 63.5 & 52.47 & 78.53 & 0.4029 \\
Harmon \citep{wu2025harmonizing} & 1.5B & 35.0 & 39.43 & 48.94 & 22.0 & 36.53 & 67.81 & 0.3917 \\
Bagel+SFT & -- &  52.6 & 80.88 & 82.65 & 73.1 & 68.27 & 78.03 &  0.2274\\
Bagel+UniCot \citep{qin2026uni} & -- & 53.1 & 80.99 & 83.12 & 83.0 & 70.00 & 77.91 & 0.4037 \\
Bagel+IRG \citep{huang2026interleaving} & -- & 48.0 & 57.35 & 60.47 & 68.0 & 61.53 & 72.06 & 0.3842 \\
\rowcolor{blue!8}Bagel+\method & -- & \textbf{54.0} & \textbf{85.87} & \textbf{86.75} & \textbf{73.4} & \textbf{70.93} & \textbf{82.03} & 0.4292 \\
\bottomrule
\end{tabular}}
\end{table}

\textbf{One model serves multiple budgets.}
The inference-time threshold $\eta_\beta$ traces an accuracy-compute curve without retraining. \Cref{tab:budget} shows that adaptive halting consistently outperforms static truncation across the operating range, with the optimal depth varying across benchmarks. This supports input- and benchmark-dependent computation allocation rather than a single fixed depth.

\begin{table}[h]
\centering
\caption{Effect of the halting threshold $\eta_\beta$ on Bagel.}
\label{tab:budget}
\small
\begin{tabular}{lrrrrrr}
\toprule
Setting & $\eta_\beta$ & $\bar d$ & Rel. compute & MMMU & MMStar & MathVista \\
\midrule
Aggressive   & 0.30 & 26.072 & 0.930 & 52.9 & 71.2 & 69.1 \\
Balanced     & 0.50 & 27.841 & 0.993 & 54.0 & 70.9 & 73.4 \\
Conservative & 0.80 & 28.000 & 1.000 & 53.6 & 70.8 & 73.1 \\
\bottomrule
\end{tabular}
\end{table}

\begin{figure*}[t]
\centering
\begin{subfigure}[t]{0.49\textwidth}
\centering
\includegraphics[width=\linewidth]{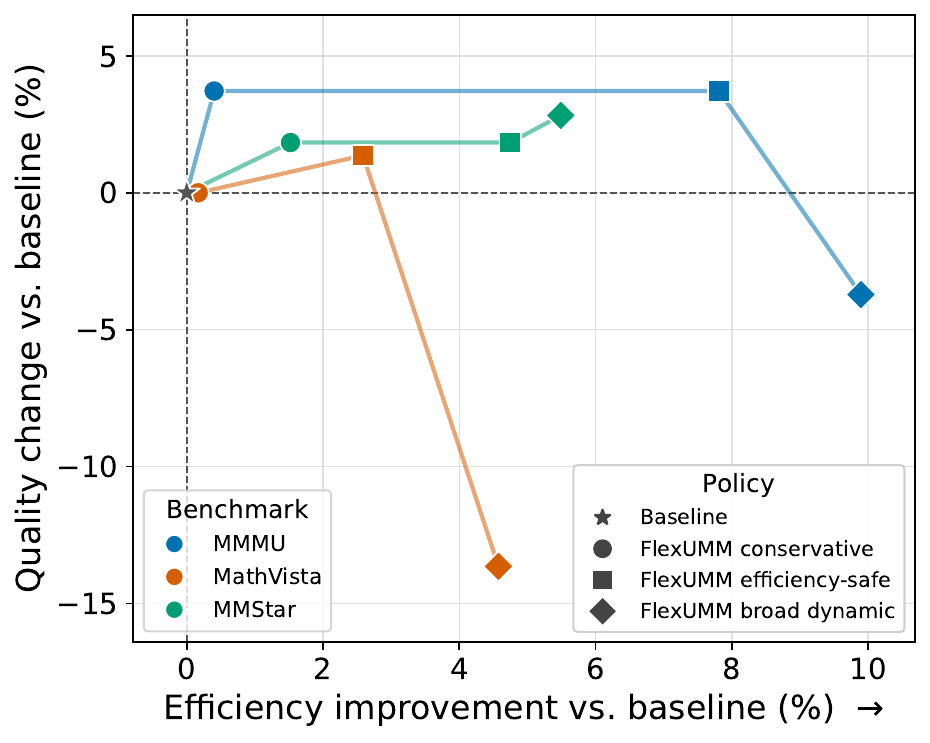}
\caption{Quality--efficiency trade-off.}
\label{fig:efficiency-quality-tradeoff}
\end{subfigure}\hfill
\begin{subfigure}[t]{0.49\textwidth}
\centering
\includegraphics[width=\linewidth]{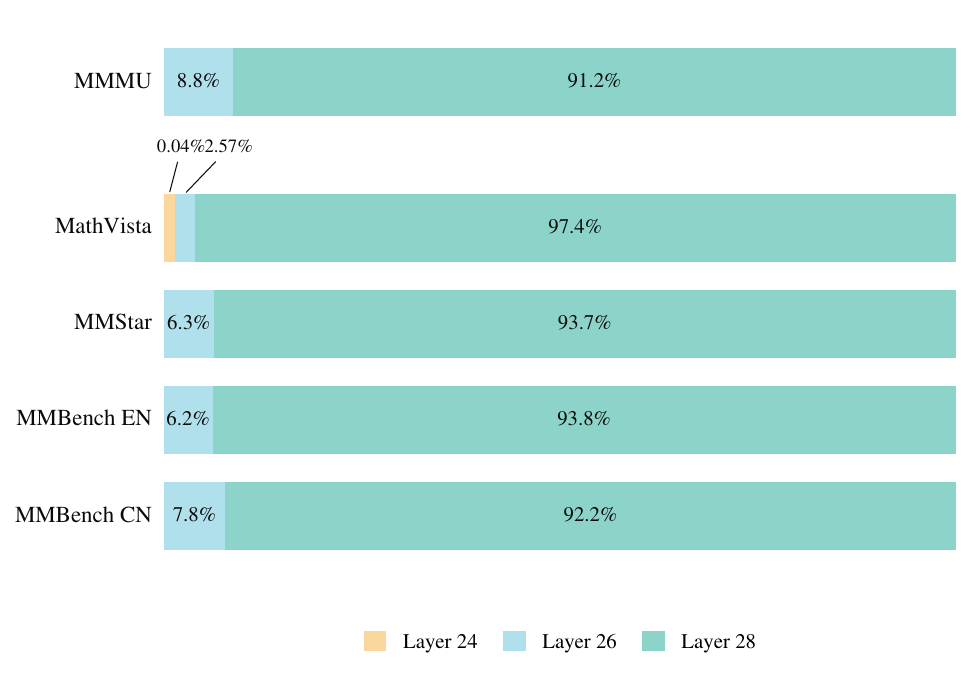}
\caption{Exit-layer distribution.}
\label{fig:exit-distribution}
\end{subfigure}
\caption{Quality-efficiency trade-off and adaptive exit distribution.}
\end{figure*}

\Cref{fig:efficiency-quality-tradeoff} further shows the accuracy-efficiency trade-off: the efficiency-safe policy preserves full-depth quality while reducing the layer-token compute proxy, whereas more aggressive early exit yields larger savings at some quality cost.

\textbf{Low adaptation overhead.}
The added routers, halting heads, and low-rank residual banks increase the checkpoint size by only 0.381\% (Appendix~\Cref{tab:hparams}).
Early exit reduces the layer-token compute proxy, while Decouple activates only a selected low-rank residual branch. 
We report these as compute proxy savings rather than direct wall-clock speedups, since the current dense implementation still evaluates the base FFN (\Cref{sec:analysis}).

\subsection{Ablation Study}
\label{sec:ablation}

Our ablations are run on Bagel in \Cref{tab:ablation}.

First, \method specializes FFN capacity while keeping attention shared, following the gradient-alignment analysis in \Cref{sec:sharing-analysis}. The first block of \Cref{tab:ablation} confirms this choice: routing attention alone underperforms the backbone on most understanding benchmarks, while routing both sublayers recovers only part of the gain. Specialization is therefore most effective when applied to the FFN while preserving shared attention.

Second, \method uses input-adaptive depth rather than a fixed truncation. 
The second block of \Cref{tab:ablation} shows that fixed depths around \method's mean exit depth are consistently suboptimal: changing the fixed depth shifts which benchmarks degrade without eliminating the loss. 
This supports the need for input-dependent depth allocation.

Third, we test whether gains come from learned routing rather than added parameters. The last block of \Cref{tab:ablation} keeps the architecture and parameter count fixed but replaces learned routing with random decisions. 
Random routing loses \method's gains and falls below the backbone on several benchmarks, confirming that performance depends on effective computation allocation rather than parameter addition alone.
The remaining design-choice ablations, including the capacity and stopping
variants, are reported in Appendix~\ref{sec:ablation-design}.

\begin{table}[t!]
\centering
\caption{Ablation study on Bagel.}
\label{tab:ablation}
\small
\setlength{\tabcolsep}{4pt}
\resizebox{\linewidth}{!}{
\begin{tabular}{lccccc|cc}
\toprule
\multirow{2}{*}{Variant} & \multicolumn{5}{c|}{Understanding} & \multicolumn{2}{c}{Generation} \\ \cline{2-6} \cline{7-8}
 & MMMU & MMBench-CN & MMBench-EN & MathVista & MMStar & GenEval & WISE \\ \midrule
Bagel & 51.9 & 80.94 & 82.65 & 71.3 & 63.20 & 78.81 & 0.4044 \\
\rowcolor{blue!8}Bagel+\method & \textbf{53.6} & \textbf{85.87} & \textbf{86.81} & \textbf{73.1} & \textbf{70.80} & \textbf{82.03} & \textbf{0.4292} \\ \midrule
\multicolumn{8}{l}{\emph{Capacity axis: which components are routed}} \\
\quad Routed attention, shared FFN & 52.1 & 82.26 & 81.95 & 71.2 & 62.30 & 81.21 & 0.4028 \\
\quad Routed attention, routed FFN & 53.2 & 84.81 & 84.24 & 71.9 & 68.90 & 80.65 & 0.3995 \\ \midrule
\multicolumn{8}{l}{\emph{Depth axis: fixed instead of input-dependent}} \\
\quad Fixed depth $d=25$ & 51.3 & 81.61 & 83.58 & 48.2 & 68.80 & 77.19 & 0.3974 \\
\quad Fixed depth $d=26$ & 52.1 & 81.01 & 84.13 & 67.2 & 67.90 & 76.87 & 0.4039 \\ \midrule
\multicolumn{8}{l}{\emph{Routing policy}} \\
\quad Random router & 52.4 & 80.48 & 74.34 & 71.0 & 64.53 & 76.38 & 0.3849 \\
\bottomrule
\end{tabular}}
\end{table}

\begin{table}[t!]
\centering
\caption{Unification test on Unison \citep{liu2026unison} and Unified-Bench \citep{yan2026unified}.}
\label{tab:unification}
\small
\setlength{\tabcolsep}{4pt}
\resizebox{\linewidth}{!}{
\begin{tabular}{lccccc|ccccc}
\toprule
\multirow{2}{*}{Method} & \multicolumn{5}{c|}{Unison} & \multicolumn{5}{c}{Unified-Bench} \\ \cline{2-6} \cline{7-11}
 & Int. Cons. & U.-G. Gen. & G.-G. Und. & Mut. Enh. & Overall & CLIP & DINOv2 & DINOv3 & LongCLIP & Overall \\ \midrule
Bagel & 80.72 & 72.09 & 33.68 & 35.87 & 55.59 & 89.47 & 78.77 & 72.40 & 93.21 & 83.46 \\
\rowcolor{blue!8}+\method & 83.21\impr{3.1} & 72.96\impr{1.2} & 37.08\impr{10.1} & 46.08\impr{28.5} & 59.83\impr{7.6} & 90.91\impr{1.6} & 79.88\impr{1.4} & 74.53\impr{2.9} & 94.87\impr{1.8} & 85.05\impr{1.9} \\
\bottomrule
\end{tabular}}
\end{table}

\subsection{Evaluation on UMM Unification}
\label{sec:unification}

We evaluate \method on two unified-model benchmarks, Unison \citep{liu2026unison} and Unified-Bench \citep{yan2026unified}. 
Effective unification requires a model to support both understanding and generation
while maintaining consistent performance across these complementary capabilities. 
\Cref{tab:unification} shows consistent improvements over the baseline on both unification benchmarks, with gains across all evaluated dimensions and overall scores. 
These results suggest that adaptive computation and input-dependent depth selection help allocate computation more flexibly across different capabilities, improving the balance between understanding and generation. 
Overall, \method enhances the unified performance of the model across both evaluated benchmarks.

\subsection{Evaluation on Safety}
\label{sec:safety}

\begin{wraptable}{r}{0.5\textwidth}
\centering
\caption{Uni-SafeBench safety results.}
\label{tab:unisafebench}
\small
\setlength{\tabcolsep}{7pt}
\resizebox{\linewidth}{!}{
\begin{tabular}{lcccc}
\toprule
Model & C-Unsafe $\downarrow$ & C-Reject $\uparrow$ & I-Unsafe $\downarrow$ & Both-Safe $\uparrow$ \\
\midrule
Bagel & 43.72\% & 5.07\% & 49.20\% & 43.96\% \\
\rowcolor{blue!8}
+\method & \textbf{42.35\%} & \textbf{5.23\%} & \textbf{49.06\%} & \textbf{44.86\%} \\
\bottomrule
\end{tabular}
}
\end{wraptable}

We further assess safety using Uni-SafeBench \citep{peng2026does}, which is important for UMMs that combine multimodal understanding with generation, requiring both accurate safety recognition and appropriate response behavior. 
As shown in \Cref{tab:unisafebench}, \method lowers both unsafe rates while increasing the rejection rate and the rate of outputs that are safe under both criteria. 
These improvements suggest that adaptive depth selection provides more flexible computation for
processing safety-relevant cues and making appropriate response decisions.

\subsection{Generalization Across Backbones}
\label{sec:generalization}

\method is agnostic to UMM architectures.
\Cref{tab:generalization} evaluates \method on two additional backbones, Janus-Pro and Harmon. 
Generalization across backbones is important for assessing whether the benefits of adaptive computation extend beyond a specific architecture or sharing structure. 
\method consistently improves understanding and generation performance across all three backbones while reducing computation, demonstrating robust generalization across unified-model architectures with different sharing structures.
The results suggest that the proposed share-specialize-stop formulation can adapt to diverse backbone designs without relying on architecture-specific modifications.

\subsection{Effectiveness Analysis}
\label{sec:analysis}

We analyze the computation structure underlying \method: attention gradients remain aligned across functionalities, while FFN gradients become increasingly competitive with depth (\Cref{sec:sharing-analysis}). 
Depth requirements also vary substantially across inputs (\Cref{sec:depth-analysis}), supporting \method's capability-adaptive capacity allocation and input-adaptive computation depth.

\textbf{Exit-layer distribution is consistent across benchmarks.}
\Cref{fig:exit-distribution} shows that across benchmarks, the router systematically exits at earlier candidate depths.
Consequently, the sequence-depth component contributes a modest reduction in relative computation across benchmarks. 
This consistent behavior indicates that adaptive depth selectively allocates less computation to easier inputs, rather than relying heavily on early exit for only specific benchmarks.

\subsection{Case Study}
\label{sec:case-study}

We inspect representative outputs under the same generation prompt and seed.

\textbf{Understanding.}
The details are provided in Appendix~\Cref{sec:extended-case-study} which cover medical terminology,
geometric reasoning, person identification, counting, and color recognition.

\begin{figure*}[t!]
  \centering
  \begin{tabular}{@{}c@{}c@{}c@{}c@{}}
    \multicolumn{4}{c}{\small\emph{a photo of a chair left of a zebra}} \\
    {\scriptsize\textbf{Baseline}} & {\scriptsize\textbf{2-depth selection}} &
    {\scriptsize\textbf{3-depth selection}} & {\scriptsize\textbf{FlexUMM (4 exits)}} \\
    \includegraphics[width=.235\textwidth]{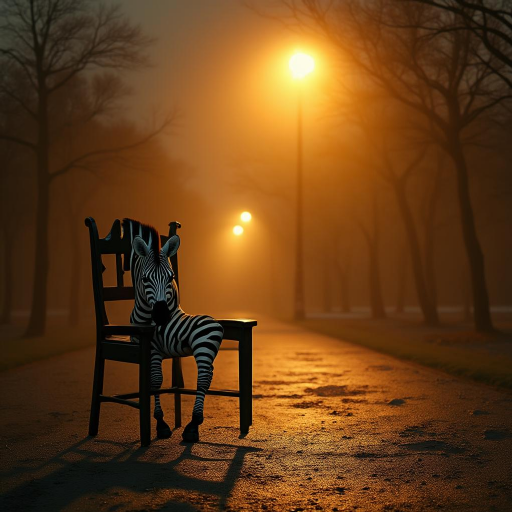} &
    \includegraphics[width=.235\textwidth]{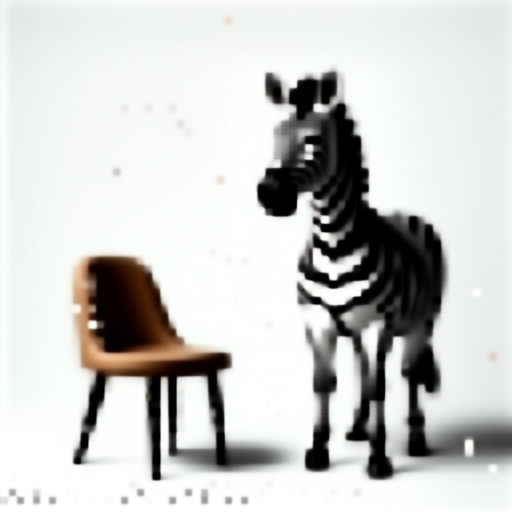} &
    \includegraphics[width=.235\textwidth]{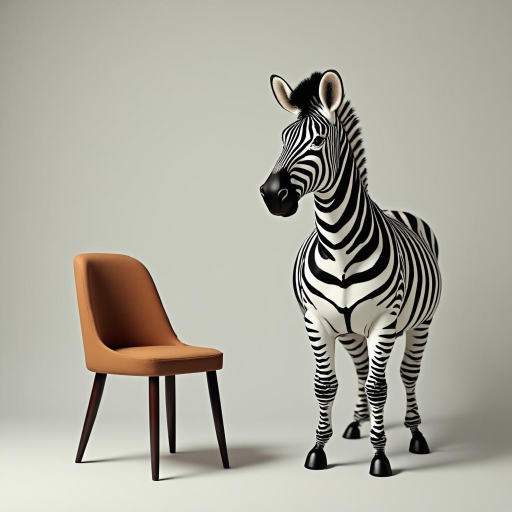} &
    \includegraphics[width=.235\textwidth]{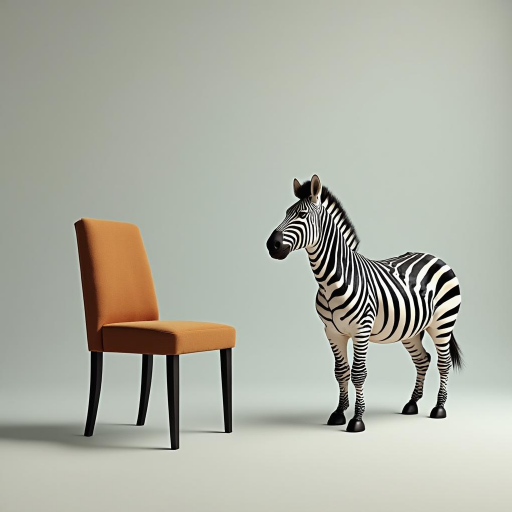}
  \end{tabular}
  \caption{Qualitative depth-selection progression for one generation prompt.}
  \label{fig:case-study-generation}
\end{figure*}

\textbf{Generation.}
We use this case study to illustrate how input-adaptive depth selection affects generation quality under matched evaluation conditions. 
All outputs are produced with the same checkpoint, prompt, random seed, and decoding settings; only the available candidate-depth configurations differ. 
The Bagel baseline executes the full backbone, whereas the 2-, 3-, and 4-exit configurations allow the policy to select from progressively finer-grained computation depths.
As shown in \Cref{fig:case-study-generation}, the baseline fails to satisfy the requested spatial relation, while adaptive depth selection progressively recovers the relation and improves visual consistency. 
The example is illustrative rather than a standalone benchmark result: Appendix~\Cref{sec:extended-case-study} reports ten strict paired understanding wins and six detector-verified generation wins, while the aggregate benchmark results quantify the overall accuracy-efficient benefit.

\begin{table}[t!]
\centering
\caption{Generalization across backbones with different sharing structures.}
\label{tab:generalization}
\small
\setlength{\tabcolsep}{4pt}
\resizebox{\linewidth}{!}{
\begin{tabular}{llccccc|cc|c}
\toprule
\multirow{2}{*}{Backbone} & \multirow{2}{*}{Method} & \multicolumn{5}{c|}{Understanding} & \multicolumn{2}{c|}{Generation} & Efficiency \\ \cline{3-7} \cline{8-9} \cline{10-10}
 &  & MMMU & MMBench-CN & MMBench-EN & MathVista & MMStar & GenEval & WISE & Rel. compute \\ \midrule
\multirow{2}{*}{Bagel} & Full backbone & 51.9 & 80.94 & 82.65 & 71.3 & 63.20 & 78.81 & 0.4044 & 1.00 \\
 & \method & 54.0\impr{4.0} & 85.87\impr{6.1} & 86.75\impr{5.0} & 73.4\impr{2.9} & 70.93\impr{12.2} & 82.03\impr{4.1} & 0.4292\impr{6.1} & 0.96 \\
\midrule
\multirow{2}{*}{Janus-Pro} & Full backbone & 40.89 & 64.84 & 69.93 & 43.8 & 47.93 & 79.02 & 0.3900 & 1.00 \\
 & \method & 46.11\impr{12.8} & 68.47\impr{5.6} & 73.11\impr{4.5} & 51.0\impr{16.4} & 52.33\impr{9.2} & 81.03\impr{2.5} & 0.4095\impr{5.0} & 0.96 \\
\midrule
\multirow{2}{*}{Harmon} & Full backbone & 35.0 & 39.43 & 48.94 & 22.0 & 36.53 & 67.81 & 0.3917 & 1.00 \\
 & \method & 38.6\impr{10.3} & 42.6\impr{8.0} & 53.09\impr{8.5} & 35.1\impr{59.5} & 43.07\impr{17.9} & 71.71\impr{5.8} & 0.4197\impr{7.1} & 0.89 \\
\bottomrule
\end{tabular}}
\end{table}

\section{Conclusion}

We study adaptive computation in unified multimodal models and propose \method, a unified framework that dynamically allocates computation according to both capability and input complexity. 
By adapting computation within and across layers while preserving a shared multimodal backbone, \method enables the model to allocate computation where it is most useful rather than applying uniform computation to all tokens and inputs. 
Across three unified multimodal backbones, \method consistently improves overall performance, achieving an
average improvement of \textbf{10.17\%} while reducing FLOPs by \textbf{6.33\%}.
These results show that adaptive computation provides an effective principle for improving both the capability and efficiency of unified multimodal models.
We discuss limitations in Appendix~\ref{sec:limitations}.

\subsection*{AI use statement}
We used generative AI tools to assist with manuscript revision, literature search, and code development. The experimental design and methodological choices were made by the authors.

\subsection*{Reproducibility statement}
Dataset protocols, preprocessing, model pools, and training hyperparameters are provided in Appendix~\ref{sec:configuration}.

\bibliography{references}
\bibliographystyle{plainnat}

\appendix
\onecolumn

\begin{center}
    {\large \textbf{Appendix\\
    \method: FlexUMM}}
\end{center}

\etocdepthtag.toc{mtappendix}
\etocsettagdepth{mtchapter}{none}
\etocsettagdepth{mtappendix}{subsection}
\tableofcontents

\clearpage

\section{Configuration}
\label{sec:configuration}

This section describes the runtime configuration of \method.  It focuses on
how the gates, residual banks, and depth controller are inserted into the
shared decoder and how they interact during a forward pass.  It intentionally
does not describe parameter selection, tuning schedules, or the post-training
procedure.

\subsection{Backbone configurations}
\label{sec:backbone-configurations}

The backbone settings and the number of trained parameters used in our
experiments are summarized in \Cref{tab:hparams}.

\begin{table}[h]
\centering
\caption{Backbone configurations for \method.}
\label{tab:hparams}
\small
\setlength{\tabcolsep}{5pt}
\begin{tabular}{lccccc}
\toprule
Backbone & Params & Lang. layers & Candidate depths $\mathcal{D}$ & Experts $E$ & Trained params \\
\midrule
Bagel & 14.607B & 28 & $\{22,24,26,28\}$ & 2 & 0.056B (+0.381\%) \\
Janus-Pro & 7.420B & 30 & $\{21,24,27,30\}$ & 2 & 0.035B (+0.472\%) \\
Harmon & 1.5B & 28 & $\{22,24,26,28\}$ & 2 & 0.013B (+0.837\%) \\
\bottomrule
\end{tabular}
\end{table}

\subsection{Insertion into the shared decoder}

The runtime configuration attaches \method to the language-decoder stack
of the UMM rather than constructing a second model.  Each decoder layer keeps
its pretrained attention, normalization, and base FFN modules.  For a layer
$\ell$, the shared part of the computation is therefore the ordinary decoder
operation
\begin{equation}
    \tilde{\vh}^{(\ell)}
    = \vh^{(\ell-1)}
      + \operatorname{Attn}^{(\ell)}
        \!\left(\operatorname{Norm}_1^{(\ell)}
        \!\left(\vh^{(\ell-1)}\right)\right),
    \qquad
    \vu^{(\ell)}
    = \operatorname{Norm}_2^{(\ell)}
        \!\left(\tilde{\vh}^{(\ell)}\right).
\end{equation}
The TAFE module is placed at the FFN call.  Thus, attention remains the
shared communication channel, while the FFN residual is the component whose
capacity can be shared, specialized, or suppressed.  Understanding and
generation have separate TAFE controllers and residual banks, even when both
token streams occur in one packed transformer forward.

The runtime preserves the original names and parameter objects of the
base FFN projections.  A checkpoint can therefore load the pretrained
attention, normalization, and FFN weights directly; \method adds only the
controller modules and residual banks around those weights.  Packed batches
carry token indexes, sample lengths, and task metadata.  Before a gate is
called, the runtime maps each routed token back to its packed sample so that
the correct task embedding and modality-specific controller are used.

\subsection{TAFE controller}

For token $k$ at layer $\ell$, the gate forms a route feature by combining the
current hidden state with task, layer, route-kind, and remaining-budget
information:
\begin{equation}
    \vz_k^{(\ell)}
    = W_s\,\operatorname{LN}(\vu_k^{(\ell)})
      + \ve_{\mathrm{task}(k)}
      + \ve_{\mathrm{layer}(\ell)}
      + \ve_{\mathrm{kind}}
      + W_b b^{(\ell)}_k.
\end{equation}
Two small MLP heads read this feature.  The utility head produces three
values, one for each action in
$\{\textsc{Share},\textsc{Decouple},\textsc{Exit}\}$, and the subset head
produces one logit for each residual bank.  The action decision uses the
cost-adjusted utility in Eq.~(\ref{eq:gate_prob}); the subset decision is the
argmax bank index (or the top-$k$ bank indexes when multiple residuals are
enabled).  The subset index is predicted from the hidden state and metadata;
it is not a human-supplied task label.

The hard inference path is implemented tokenwise.  For a \textsc{Share} token,
the base FFN output is retained.  For a \textsc{Decouple} token, one selected
residual bank is evaluated and added to the base output.  For an \textsc{Exit}
token, the FFN output is replaced by zero, so the decoder's outer residual
connection carries the post-attention state forward unchanged by the FFN.
If top-$k$ routing is used, the selected residual outputs are averaged before
being added.  The resulting FFN update is exactly the piecewise operation in
Eq.~(\ref{eq:capacity_update}).

During differentiable routing, the hard action is replaced by the soft
mixture $\pi$ from Eq.~(\ref{eq:gate_prob}).  With $\rho_{k,e}$ denoting the
soft bank weights, the routed FFN output is
\begin{equation}
    \Delta\vh_k^{(\ell)}
    = \left(\pi_{k,\textsc{Share}}^{(\ell)}
      + \pi_{k,\textsc{Decouple}}^{(\ell)}\right)F(\vu_k^{(\ell)})
      + \pi_{k,\textsc{Decouple}}^{(\ell)}
        \sum_{e\in\mathcal{T}_k}\rho_{k,e}^{(\ell)}
        R_e(\vu_k^{(\ell)}),
\end{equation}
where $\mathcal{T}_k$ is the selected top-$k$ bank set.  The \textsc{Exit}
probability contributes neither the shared FFN output nor a specialized
residual.  Evaluation switches back to the discrete argmax route.

\subsection{Low-rank residual banks}

Each configured specialized bank is a residual adapter rather than a copied
FFN block.
For bank $e$, the configured residual block computes
\begin{equation}
    R_e(\vu)
    = U_e\,\operatorname{SiLU}\!\left(D_e\,\operatorname{RMSNorm}(\vu)\right),
    \qquad
    D_e\in\mathbb{R}^{r\times d},\quad
    U_e\in\mathbb{R}^{d\times r}.
\end{equation}
The down projection is initialized with a small random matrix and the up
projection is initialized to zero.  Consequently, inserting the banks does
not change the released backbone's initial forward function, while each bank
can learn a different functionality- or modality-conditioned residual.  The
dense wrapper first evaluates the pretrained base FFN $F(\vu)$ for every
routed token.  The selected action is then applied to that result: a
\textsc{Share} token retains $F(\vu)$, a \textsc{Decouple} token adds one
selected low-rank residual $R_e(\vu)$, and an \textsc{Exit} token masks the
FFN update after the base result has been formed.  Thus \textsc{Decouple}
adds only a low-rank residual bank (roughly the two rank-$r$ projections)
rather than a second full FFN, but \textsc{Share} or \textsc{Exit} should
not be interpreted as removing the dense base-FFN kernel in this runtime.
The action-level compute proxy credits the low-rank branch only to selected
tokens; direct execution savings come from terminating the sequence before
later decoder layers.

\subsection{Candidate-depth execution}

The depth controller is separate from the tokenwise TAFE gate.  It is a small
MLP shared across the configured candidate exits.  At candidate depth $d_m$,
the controller pools the current shared hidden state for the routed sequence
and combines that pooled state with the candidate-depth embedding, route-kind
embedding, task embedding, and generation timestep when one is available.  An
optional previous-candidate state supplies a representation-change feature.
The controller returns a halt logit and, when enabled, a predicted harmful
regret and uncertainty scale.

The runtime evaluates layers only up to the largest configured candidate.  At
each candidate it computes $p_{\mathrm{halt},m}$ and applies the eligibility threshold from
Eq.~(\ref{eq:inference_exit}).  The first eligible candidate is selected; the
last candidate is always the fallback.  If all active token routes take
\textsc{Exit} at an earlier layer, the decoder loop can terminate immediately,
before the next candidate-depth check.  Thus the two controls have different
granularities: TAFE acts on token--layer pairs, while the depth controller
terminates the remaining sequence-level computation.

\subsubsection{Quality-constrained router training}
\label{sec:router-training-details}

The depth router is trained from all configured candidate depths in one forward
pass.  If the candidate depths are $d_1<\cdots<d_M$, the runtime stores the
hidden state $h_{i,m}$ for sample $i$ at every $d_m$ while executing the
backbone to $d_M$.  It then applies the existing task head to every candidate:
the language-model head produces a teacher-forced cross-entropy loss for
understanding, while the flow-to-VAE head produces a flow-matching MSE against
the same generation target used at the final depth.  Hence the router is not
trained from a single early-exit head or from a sampled candidate; every
candidate contributes a task-cost value $c_{i,m}$ on every training example.

The quality constraint is defined relative to the final candidate.  Let
$\delta$ be the configured quality margin and let $a_{i,m}$ denote any
task-specific validity mask (for example, the label-correct target used by the
understanding router).  The candidate is eligible when
\begin{equation}
    s_{i,m}
    = \mathbf{1}\!\left[c_{i,m}\leq c_{i,M}+\delta\right]a_{i,m}=1,
    \qquad s_{i,M}=1.
    \label{eq:quality-eligibility}
\end{equation}
The final-candidate override guarantees a valid target even when no earlier
candidate satisfies the quality constraint.  The default supervised target is
the first eligible candidate,
\begin{equation}
    t_i=\min\{m:s_{i,m}=1\},
    \label{eq:earliest-safe-target}
\end{equation}
although the implementation also supports selecting the lowest-cost or best
candidate target when that strategy is requested.  The candidate costs,
eligibility masks, and teacher-regret labels are detached before they supervise
the router; they are labels for allocation, not an additional gradient path
through the router into the backbone.

The sequential probabilities $q_{i,m}$ in Eq.~(\ref{eq:exit_prob}) are the
differentiable training surrogate for the first-halt policy.  The router-side
objective combines the expected detached candidate cost, a normalized expected
depth, entropy regularization, and supervised/calibration terms:
\begin{equation}
\begin{split}
    \mathcal{L}_{\mathrm{router}}
    ={}&\lambda_{\mathrm{task}}
      \sum_m q_{i,m}\,\operatorname{sg}(c_{i,m})
      +\lambda_{\mathrm{depth}}\sum_m q_{i,m}\frac{d_m}{d_M}
      -\lambda_{\mathrm{ent}}\,H(q_i) \\
     &+\lambda_{\mathrm{sup}}\mathcal{L}_{\mathrm{sup}}(q_i,t_i)
      +\lambda_{\mathrm{cal}}\mathcal{L}_{\mathrm{cal}}(q_i,s_i)
      +\lambda_{\mathrm{risk}}\mathcal{L}_{\mathrm{risk}}.
\end{split}
\label{eq:router-training-objective}
\end{equation}
Here $\operatorname{sg}$ denotes stop-gradient.  The supervised term is a
masked binary-cross-entropy loss on the sequential halt hazards: candidates
before $t_i$ are trained to continue, candidate $t_i$ to halt, and later
candidates are not treated as competing positive exits.  The calibration term
aligns the predicted halt behavior with the eligible-candidate mask.  The
optional risk head predicts both a harmful-regret probability $\hat r_{i,m}$
and a positive uncertainty scale $\hat u_{i,m}$.  Its loss is a binary
cross-entropy term for the task-specific risk target plus a weighted smooth-$L_1$
term that calibrates the uncertainty scale.  For understanding, the harmful-
regret target marks a candidate only when it is wrong while the final-candidate
teacher is correct, so correcting a teacher mistake is not penalized.  For
generation, the risk target is built from normalized excess flow-MSE relative
to the final candidate (together with the configured candidate-cost strategy),
rather than from label correctness.

For clarity, the complete training sequence is: (1) run to the deepest
candidate and collect every candidate hidden state; (2) evaluate every native
task head and form the per-candidate task costs; (3) construct quality,
label-correctness, and harmful-regret targets from detached candidate and final
candidate predictions; (4) form $q_{i,m}$ and the router objective above; and
(5) backpropagate the native task loss, multi-exit task losses, final-depth
hidden-state distillation, router loss, and TAFE loss through the configured
trainable adaptation modules.  The router input features are detached as well,
so the router cannot reshape the backbone representation merely to satisfy its
allocation labels.

At inference, the router scans candidates in increasing depth.  It halts at
the first candidate satisfying the minimum-depth condition, the halt
probability threshold, and when the risk head is enabled, the upper confidence
bound
\begin{equation}
    \hat r_{i,m}+\kappa\hat u_{i,m}\leq\tau_{\mathrm{risk}}.
    \label{eq:risk-upper-bound}
\end{equation}
The final candidate remains an unconditional fallback.  Thus $q_{i,m}$ is a
smooth surrogate used during training, whereas deployment uses the first
eligible candidate with fixed validation-time thresholds; it does not inspect
test labels or recompute the quality constraint at inference.

\subsection{Forward-pass sequence}

For one packed forward pass, the configured computation proceeds as follows:
\begin{enumerate}[leftmargin=*]
    \item The runtime identifies the understanding and generation token
    streams, records their packed positions, and associates each sample with
    its route kind and task metadata.
    \item At each decoder layer, the shared attention and normalization run on
    the packed sequence.  The corresponding TAFE gate is then called on each
    active stream, and the selected FFN residual is written back to the same
    packed positions.
    \item After the layer update, an all-route \textsc{Exit} condition may
    stop the decoder loop.  Otherwise, at a configured candidate depth, the
    depth controller decides whether the sequence continues.
    \item The selected hidden state is passed to the backbone's existing
    output head: the language-model head for understanding or the
    flow-to-VAE head for generation.  No independent prediction head is
    introduced for an intermediate depth.
\end{enumerate}

\subsection{Metrics}
\label{sec:metrics}

We report three efficiency quantities.  \emph{Relative compute} is a normalized
compute proxy, $\E[\mathcal{C}_{\mathrm{proxy}}]/\mathcal{C}_{\max}$, where the
reference is the same backbone run fully shared to full depth.  The proxy has
two components: (i) layer--token work, which decreases when the depth policy
stops a sequence and skips the remaining layers, and (ii) the action-conditioned
FFN term, which charges a full shared-FFN unit to \textsc{Share} and only the
selected low-rank residual-bank work to \textsc{Decouple}.  In the routing
telemetry this action term uses nominal relative costs of $1.0$, $0.75$, and
$0.0$ for \textsc{Share}, \textsc{Decouple}, and \textsc{Exit}, respectively;
these weights describe allocation, not hardware-measured FLOPs.  The latter is
important because the dense base FFN is evaluated before the tokenwise action
mask in the current wrapper.  Thus the direct and robust runtime saving is the
sequence-level skip; the realized latency benefit of the low-rank term depends
on sparse or fused kernel support and is not measured by this paper's proxy.
Consequently, the relative values in Table~\ref{tab:generalization} should not
be read as early-exit percentages or as wall-clock speedups: they combine the
depth and action-level accounting terms, while launch, packing, and memory
effects are outside the proxy.
\emph{Mean exit depth} $\bar{d} = \sum_{\ell \in \mathcal{D}} q^{(\ell)}\ell$ summarizes when the model stops computing.
\emph{Specialization rate} $\bar{\alpha} = \frac{1}{NL}\sum_{i=1}^{N}\sum_{\ell=1}^{L} \mathbf{1}\big[a_i^{(\ell)} = \mathrm{DECOUPLE}\big]$ is the fraction of the $N{\times}L$ token--layer pairs that take the \textsc{Decouple} action of \Cref{eq:capacity_update}, and summarizes how much capacity specializes.

\section{Findings}

\subsection{Sharing Is Beneficial Early, But Not Uniformly Throughout Depth}
\label{sec:sharing-analysis}

In early layers, different functionalities can benefit from processing a common representation. Shared attention provides a mechanism for information exchange across modalities and functionalities, allowing different pathways to use complementary contextual information. However, as representations become increasingly task-specific, forcing all functionalities through the same representational capacity can induce competition. This is particularly problematic for the FFN, which provides much of the transformer's representational capacity.

\subsubsection{Controlled (Synthetic) Layerwise Competition Diagnostic}

This study is designed to answer one mechanistic question: when two functionalities share the same transformer component, do their learning signals update that component in compatible or conflicting directions? We use a synthetic two-task setting so that the amount of shared structure and task-private pressure can be controlled directly.  The shared target, private targets, and depth-increasing private weight are deliberately constructed interventions.  They are useful for isolating a mechanism, but they are not measurements of a released Bagel, Janus-Pro, Harmon, or Show-o backbone.

Each input sequence is associated with two tasks, denoted task $A$ and task $B$. Both tasks depend on a shared target $y^{\mathrm{sh}}$, but each task also has its own private target, $y^{A}$ or $y^{B}$. The shared target creates a reason for the two tasks to benefit from common computation, while the private targets create task-specific pressure. For layer $\ell$, we write the task losses as
\begin{align}
    \mathcal{L}^{A}_{\ell}
    &= \mathcal{L}^{\mathrm{sh}}_{\ell} + \lambda_{\ell}\mathcal{L}^{A,\mathrm{priv}}_{\ell}, \\
    \mathcal{L}^{B}_{\ell}
    &= \mathcal{L}^{\mathrm{sh}}_{\ell} + \lambda_{\ell}\mathcal{L}^{B,\mathrm{priv}}_{\ell}.
\end{align}
Here, $\mathcal{L}^{\mathrm{sh}}_{\ell}$ is the shared objective at layer $\ell$, $\mathcal{L}^{A,\mathrm{priv}}_{\ell}$ and $\mathcal{L}^{B,\mathrm{priv}}_{\ell}$ are task-private objectives, and $\lambda_{\ell}$ controls the strength of the private objective at depth $\ell$. Increasing $\lambda_{\ell}$ with depth lets us isolate how growing task-specific demands affect shared computation.

For a transformer component $c$ at layer $\ell$, such as attention, normalization, or FFN, let $\theta^{c}_{\ell}$ denote its parameters. We compute the task-specific gradients
\begin{align}
    g^{A,c}_{\ell} &= \nabla_{\theta^{c}_{\ell}}\mathcal{L}^{A}_{\ell}, \\
    g^{B,c}_{\ell} &= \nabla_{\theta^{c}_{\ell}}\mathcal{L}^{B}_{\ell}.
\end{align}
The alignment between the two tasks for component $c$ at layer $\ell$ is measured by gradient cosine similarity:
\begin{equation}
    \operatorname{cos}^{c}_{\ell}
    =
    \frac{\langle g^{A,c}_{\ell}, g^{B,c}_{\ell}\rangle}
    {\|g^{A,c}_{\ell}\|_2\|g^{B,c}_{\ell}\|_2}.
\end{equation}
Positive cosine means the two tasks update the shared component in compatible directions. Negative cosine means the tasks compete: improving one task pushes the shared component in a direction that hurts the other. We report FFN competition as
\begin{equation}
    \operatorname{comp}^{\mathrm{FFN}}_{\ell}
    =
    \max(0, -\operatorname{cos}^{\mathrm{FFN}}_{\ell}).
\end{equation}
Thus, a competition value of 0 means no negative alignment, while larger values indicate stronger conflict.

For a model with $L$ layers, the depth-dependent private weight is
\begin{equation}
    \lambda_{\ell}
    =
    0.15 + 2.5
    \left(\frac{\ell - 1}{L - 1}\right)^{1.35},
    \qquad \ell \in \{1,\ldots,L\}.
\end{equation}
Thus, $\lambda_1=0.15$ in the first layer and $\lambda_L=2.65$ in the final layer for every model depth. The three regimes use the same $\lambda_{\ell}$ schedule and differ by the private-target amplitude $\alpha$, which scales the opposing task-private target before it enters $\mathcal{L}^{A,\mathrm{priv}}_{\ell}$ and $\mathcal{L}^{B,\mathrm{priv}}_{\ell}$. For readability, Table~\ref{tab:private-pressure-settings} also reports the endpoint range of the amplitude-weighted private pressure, $\alpha\lambda_1 \to \alpha\lambda_L$.
\begin{table}[h]
\centering
\caption{Private-pressure settings for the controlled layerwise competition study.}
\label{tab:private-pressure-settings}
\small
\begin{tabular}{llll}
\toprule
Regime & $\alpha$ & $\alpha\lambda_1 \to \alpha\lambda_L$ & Description \\
\midrule
\texttt{weak-private} & 0.5 & $0.075 \to 1.325$ & Half-amplitude opposing private target \\
\texttt{medium-private} & 1.0 & $0.150 \to 2.650$ & Unit-amplitude opposing private target \\
\texttt{strong-private} & 1.5 & $0.225 \to 3.975$ & 1.5$\times$-amplitude opposing private target \\
\bottomrule
\end{tabular}
\end{table}

We evaluate three model configurations (4, 8, and 12 layers), these three private-objective regimes, eight random seeds per condition, and 16,384 gradient samples per condition. The \texttt{weak-private} regime keeps private pressure small, the \texttt{strong-private} regime makes private pressure dominant, and \texttt{medium-private} uses an intermediate setting where the tasks still share context but develop increasingly task-specific objectives with depth.

\begin{table}[h]
\centering
\caption{Layerwise gradient alignment for the 8-layer model under the \texttt{medium-private} regime.}
\label{tab:layerwise-medium-private}
\small
\begin{tabular}{lccc}
\toprule
Layer & Attention cosine & FFN cosine & FFN competition \\
\midrule
1 & 1.000 & 0.955 & 0.000 \\
2 & 1.000 & 0.796 & 0.000 \\
3 & 1.000 & 0.439 & 0.000 \\
4 & 1.000 & 0.034 & 0.004 \\
5 & 1.000 & -0.293 & 0.293 \\
6 & 1.000 & -0.520 & 0.520 \\
7 & 1.000 & -0.663 & 0.663 \\
8 & 1.000 & -0.758 & 0.758 \\
\bottomrule
\end{tabular}
\end{table}

Within this controlled family of synthetic conditions, we observe a consistent asymmetric pattern: shared attention maintains strong alignment across depth, while FFN alignment progressively deteriorates and becomes increasingly competitive in deeper layers.  This conditional result is a design hypothesis and stress test, not evidence that the same gradient trajectory must occur in every production backbone.
For example, in the 8-layer \texttt{medium-private} setting, FFN cosine drops from $0.955$ in layer 1 to $-0.758$ in layer 8, while attention cosine remains $1.000$ throughout.

\begin{table}[h]
\centering
\caption{Aggregate layerwise trends from the first to the final layer.}
\label{tab:aggregate-layerwise}
\small
\setlength{\tabcolsep}{5pt}
\begin{tabular}{llll}
\toprule
Model & Regime & Attention cosine & FFN cosine / competition \\
\midrule
4L-8H & medium-private & $1.000 \to 1.000$ & $0.954 \to -0.757$ / $0.000 \to 0.757$ \\
4L-8H & weak-private & $1.000 \to 1.000$ & $0.988 \to -0.316$ / $0.000 \to 0.316$ \\
4L-8H & strong-private & $1.000 \to 1.000$ & $0.900 \to -0.883$ / $0.000 \to 0.883$ \\
8L-8H & medium-private & $1.000 \to 1.000$ & $0.955 \to -0.758$ / $0.000 \to 0.758$ \\
8L-8H & weak-private & $1.000 \to 1.000$ & $0.989 \to -0.249$ / $0.000 \to 0.249$ \\
8L-8H & strong-private & $1.000 \to 1.000$ & $0.906 \to -0.878$ / $0.000 \to 0.878$ \\
12L-16H & medium-private & $1.000 \to 1.000$ & $0.956 \to -0.751$ / $0.000 \to 0.751$ \\
12L-16H & weak-private & $1.000 \to 1.000$ & $0.989 \to -0.287$ / $0.000 \to 0.287$ \\
12L-16H & strong-private & $1.000 \to 1.000$ & $0.903 \to -0.883$ / $0.000 \to 0.883$ \\
\bottomrule
\end{tabular}
\end{table}

\subsubsection{Shared Normalization Ablation}

We further isolate the role of shared normalization using the same controlled two-task setting. This ablation separates three possible sources of alignment: shared attention, shared normalization, and shared FFN capacity. The objective is to determine whether positive cross-task alignment comes only from attention, or whether normalization can also provide a stabilizing shared pathway.

The notation is the same as in the layerwise study. ``Shared Attn.'' indicates whether task $A$ and task $B$ use the same attention parameters $\theta^{\mathrm{attn}}_{\ell}$. ``Shared Norm'' indicates whether the tasks use the same normalization parameters $\theta^{\mathrm{norm}}_{\ell}$. When a component is private, each task has its own copy of the parameters, so there is no shared component gradient cosine to report for that component. This is marked as N/A in the table.

The FFN objective in this ablation is deliberately constructed to oppose the two tasks, so the FFN cosine is expected to be negative. This controlled design lets us ask whether shared attention and shared normalization can remain aligned even when the capacity-bearing FFN objective is maximally competitive.

\begin{table}[h]
\centering
\caption{Shared normalization ablation in the controlled two-task setting.}
\label{tab:shared-norm-ablation}
\small
\setlength{\tabcolsep}{4pt}
\resizebox{\linewidth}{!}{
\begin{tabular}{lccccc}
\toprule
Condition & Shared Attn. & Shared Norm & Attn. cosine & Norm cosine & FFN cosine \\
\midrule
Shared Attn. + Shared Norm & \cmark & \cmark & $0.989 \to 0.989$ & $1.000 \to 1.000$ & $-1.000 \to -1.000$ \\
Shared Attn. + Private Norm & \cmark & \xmark & $0.989 \to 0.989$ & N/A & $-1.000 \to -1.000$ \\
Private Attn. + Shared Norm & \xmark & \cmark & N/A & $1.000 \to 1.000$ & $-1.000 \to -1.000$ \\
Private Attn. + Private Norm & \xmark & \xmark & N/A & N/A & $-1.000 \to -1.000$ \\
\bottomrule
\end{tabular}}
\end{table}

The results reveal an important asymmetry in unified computation: communication and capacity do not have the same optimal sharing pattern. The key observation is not simply that competition increases with depth, but that different components exhibit fundamentally different sharing profiles. Attention remains aligned while FFN capacity becomes increasingly task-specific. This motivates the first design principle of \method: preserve shared communication while selectively specializing task-specific capacity.

\paragraph{Evidence boundary.}
The paper therefore separates the mechanistic diagnostic from the real-model
evidence.  The gradient cosines above are synthetic and human-designed; they
show that the proposed sharing asymmetry can arise from increasing private
pressure, but they do not establish its magnitude in a production backbone.
The empirical tests on Bagel and Janus-Pro (and the additional Harmon benchmark)
measure task quality, routing behavior, and compute proxies.  The separate
Show-o2 token-importance experiment is a real Show-o-family analysis of
functionality-specific importance, not a direct layerwise gradient-cosine
measurement.  These real-model results test whether the resulting flexible
allocation policy is useful; they should not be read as a direct validation of
the exact synthetic gradient values.  Direct component-gradient measurements
on each released backbone would be a separate experiment.

\subsubsection{Layerwise Token-Importance Sharing Across Functionalities}
\label{sec:layerwise_token_importance_sharing}

We further examine whether a fixed degree of sharing between functionalities is appropriate throughout computation in a real UMM. We use Show-o2 and estimate token importance separately for understanding and generation, then compare the resulting functionality-specific importance patterns on the same held-out inputs.

\paragraph{Token-importance estimation.}
For token $i$ at layer $\ell$, we define the layer-induced representation update as
\begin{equation}
    \Delta h_{\ell,i}
    =
    h^{\mathrm{out}}_{\ell,i}
    -
    h^{\mathrm{in}}_{\ell,i}.
\end{equation}
Following first-order Taylor importance estimation \citep{molchanov2019importance}, we assign each token the score
\begin{equation}
    s_{\ell,i}
    =
    \left|
    \nabla_{h^{\mathrm{out}}_{\ell,i}}\mathcal{L}^{\top}
    \Delta h_{\ell,i}
    \right|,
    \label{eq:gradient_update_score}
\end{equation}
which measures the alignment between the layer update and the gradient of the task loss.

For understanding, $\mathcal{L}$ is the cross-entropy loss on the expected answer tokens under teacher forcing. For generation, $\mathcal{L}$ is the flow-matching MSE loss computed along the generation trajectory using a Show-o2-generated reference latent. Show-o2 remains frozen when collecting these scores.

\paragraph{Functionality-specific predictors.}
We train two lightweight predictors, denoted MLP-U and MLP-G, to estimate the log Gradient $\times$ Update score from the token hidden state and layer identity. MLP-U is trained only on understanding samples, while MLP-G is trained only on generation samples.

The understanding set contains four task categories: counting, spatial relations, attribute combinations, and visual-state recognition, with six samples per category. For each category, four samples are used for training, one for validation, and one for testing. The generation set contains 12 samples, split into 8 for training, 2 for validation, and 2 for testing, with all 49 generation steps recorded. Each predictor is trained with two random seeds, and the evaluation results are averaged across seeds.

As a basic check that the predictors capture their corresponding importance signals, we compare their predicted top-20\% tokens with those selected directly by the Gradient $\times$ Update scores. MLP-U reaches $64.8\%$ overlap on understanding test samples, while MLP-G reaches $34.3\%$ on generation test samples. Cross-functionality transfer is lower, with $23.9\%$ for MLP-U on generation and $17.9\%$ for MLP-G on understanding. For independent random top-20\% selections, the expected normalized overlap is approximately 20\%, since each selected token has a 20\% chance of appearing in the other set.

\paragraph{Layerwise sharing.}
To compare the two functionality-specific importance patterns under the same input, we apply MLP-U and MLP-G to identical held-out samples. At each layer, each predictor selects its top $K$ tokens, where $K=\lceil 0.2N\rceil$ and $N$ is the number of valid tokens. We compute
\begin{equation}
    \mathrm{Overlap}_{\ell}
    =
    \frac{
        \left|
        \mathcal{T}^{U}_{\ell}
        \cap
        \mathcal{T}^{G}_{\ell}
        \right|
    }{K},
    \label{eq:functionality_topk_overlap}
\end{equation}
where $\mathcal{T}^{U}_{\ell}$ and $\mathcal{T}^{G}_{\ell}$ are the top-$K$ sets predicted by MLP-U and MLP-G, respectively.

\begin{figure}[t]
\centering
\includegraphics[width=0.72\linewidth]{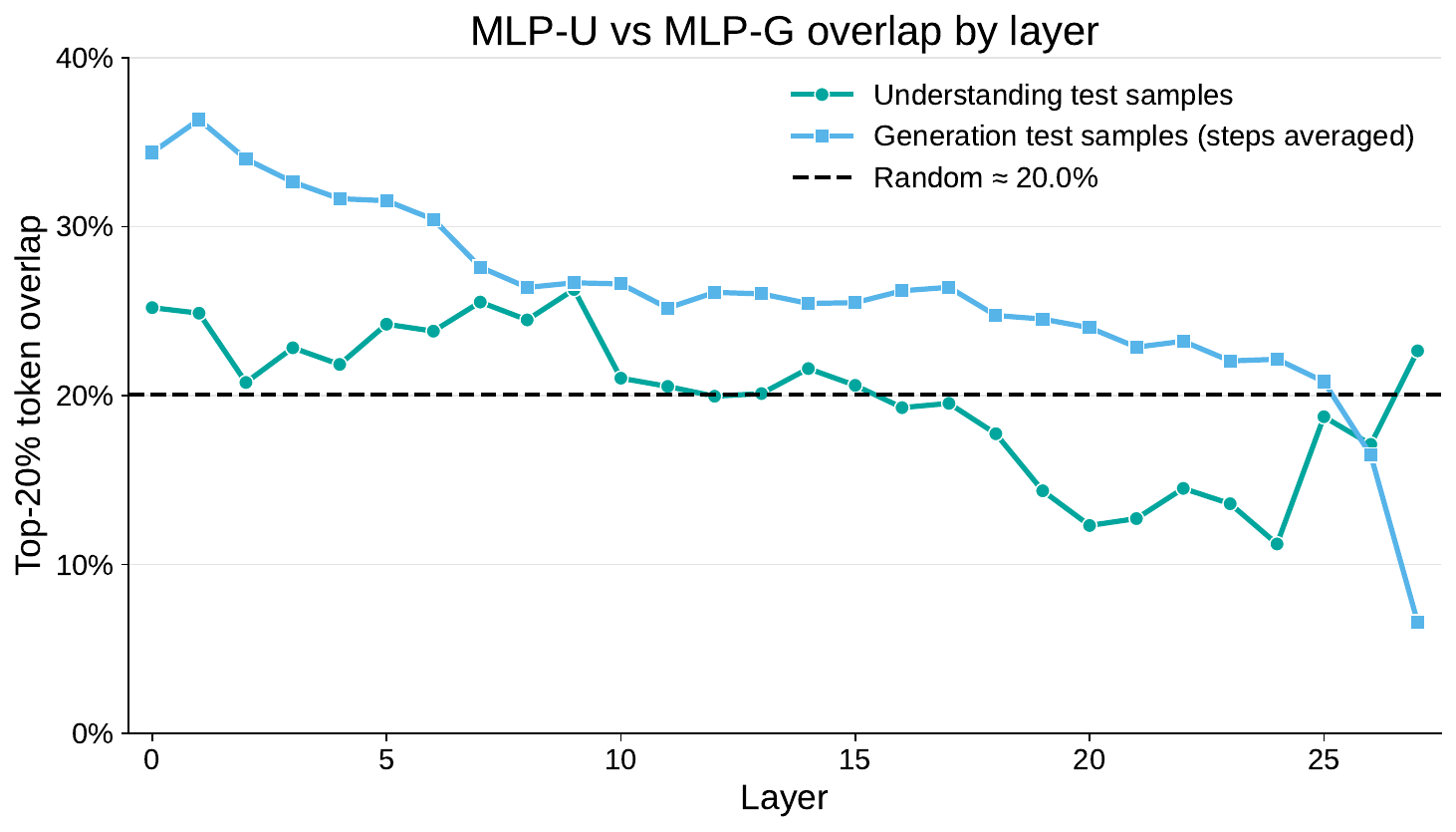}
\caption{Layerwise top-20\% token overlap between MLP-U and MLP-G predictions on held-out Show-o2 understanding and generation samples. Generation results are averaged over generation steps.}
\label{fig:ug_overlap_layers}
\end{figure}

Figure~\ref{fig:ug_overlap_layers} shows a depth-dependent pattern across both functionalities. On understanding inputs, the overlap is generally above the random baseline in earlier layers and becomes lower in deeper layers. On generation inputs, the trend is more pronounced: overlap is substantially above random in early layers and progressively decreases with depth, falling below the random baseline in the final layers. These results are consistent with stronger sharing in earlier computation and increasing functionality-specific specialization at greater depth.

\paragraph{Variation across generation steps.}
We further examine whether the relationship between the two functionality-specific predictors changes over the generation trajectory. At each generation step, we average their top-20\% token overlap across layers.

\begin{figure}[t]
\centering
\includegraphics[width=0.72\linewidth]{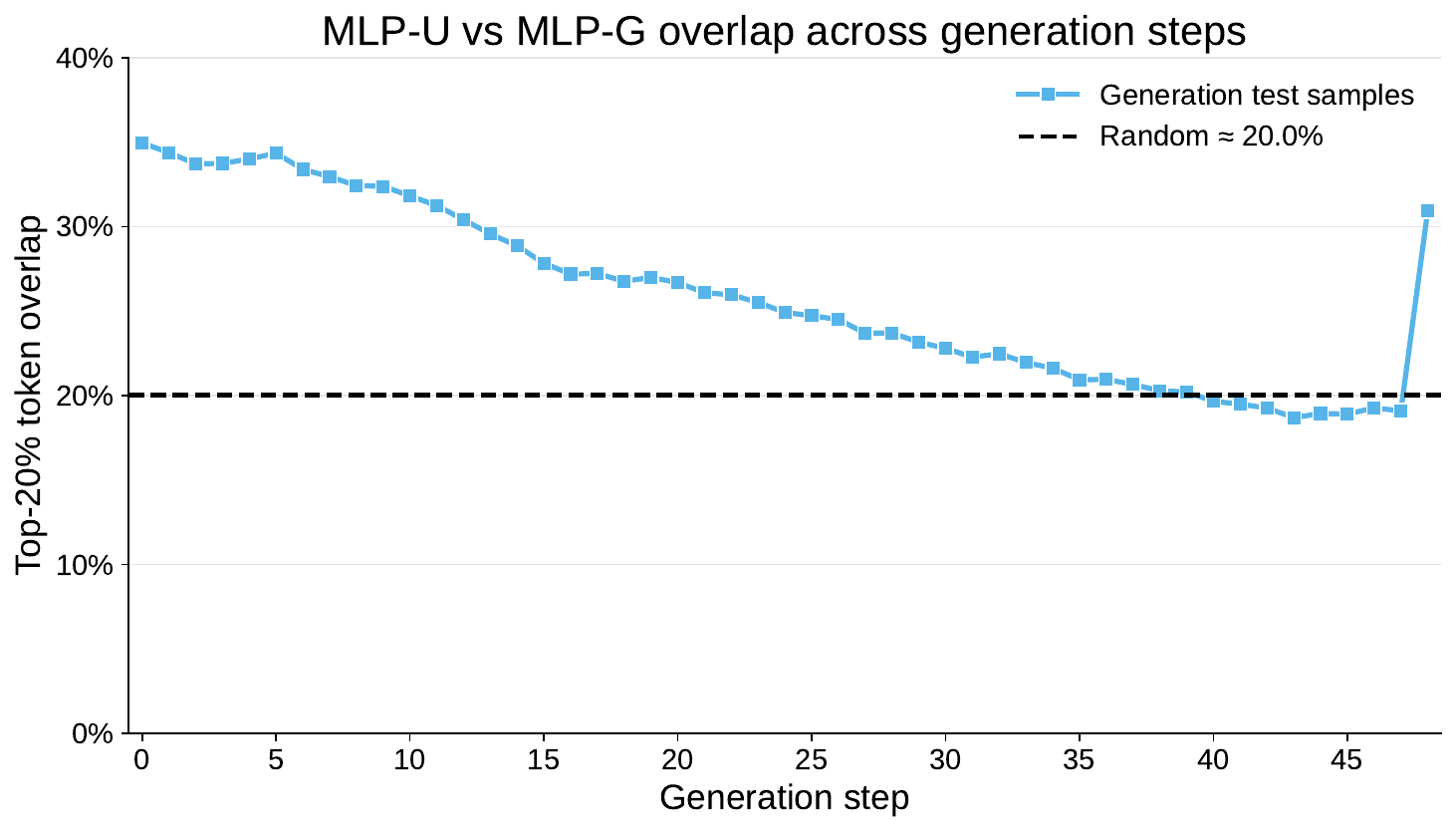}
\caption{Top-20\% token overlap between MLP-U and MLP-G predictions across generation steps on held-out Show-o2 generation samples, averaged over layers.}
\label{fig:ug_overlap_generation_steps}
\end{figure}

As shown in Figure~\ref{fig:ug_overlap_generation_steps}, the overlap generally decreases over the generation trajectory and approaches the random baseline at later steps. The final step is an exception to this overall trend. Together, the layerwise and generation-step analyses suggest that a fixed sharing pattern may be too restrictive, as functionality-specific token-importance patterns vary across both model depth and generation progress.

\subsection{Computation Depth Should Be Input-Dependent}
\label{sec:depth-analysis}

Even after accounting for parameter competition, a second limitation remains: UMMs generally require every input to traverse the same number of layers. Different inputs can reach a sufficiently mature representation at different depths. A simple understanding query may require relatively little computation, whereas a difficult reasoning problem may benefit from deeper processing. Generation inputs can also differ substantially in the amount of refinement they require.

\subsubsection{Bagel Selected-Depth Distribution on MME}

We first investigate this question using Bagel with a learned \method routing policy on the full 2,374-question MME benchmark. The goal is to test whether inputs within the same model and the same functionality prefer different amounts of computation.

Let $x_i$ denote the $i$-th MME input. Bagel contains 28 language layers, and the routing policy evaluates a candidate exit-depth set
\begin{equation}
    \mathcal{D} = \{22, 23, 25, 26, 27, 28\}.
\end{equation}
For each input $x_i$, the router selects one depth
\begin{equation}
    d_i = r_{\phi}(x_i), \qquad d_i \in \mathcal{D},
\end{equation}
where $r_{\phi}$ is the learned routing policy and $d_i$ is the selected number of language layers used for that input. A smaller $d_i$ means the model exits earlier and saves computation; a larger $d_i$ means the model continues deeper for additional refinement.

Table~\ref{tab:bagel-mme-depths} reports how often each depth is selected. ``MME samples'' is the number of benchmark examples assigned to a given depth, and ``Percentage'' is this count divided by the full 2,374-question benchmark size.

\begin{table}[h]
\centering
\caption{Selected-depth distribution for Bagel on MME.}
\label{tab:bagel-mme-depths}
\small
\begin{tabular}{lrr}
\toprule
Selected depth & MME samples & Percentage \\
\midrule
22 & 296 & 12.47\% \\
23 & 18 & 0.76\% \\
25 & 564 & 23.76\% \\
26 & 1,146 & 48.27\% \\
27 & 52 & 2.19\% \\
28 & 298 & 12.55\% \\
\midrule
Total & 2,374 & 100.00\% \\
\bottomrule
\end{tabular}
\end{table}

The selected depth ranges from 22 to 28 layers, with a mean selected depth of 25.514, a minimum of 22, a maximum of 28, and six distinct selected depths. Nearly half of the examples select depth 26, but substantial mass remains at both extremes: 12.47\% exit at depth 22 and 12.55\% run the full 28 layers. This variation occurs within a single model and a single benchmark, rather than simply reflecting differences between understanding and generation tasks; it is genuine input-level variation in how much computation an example requires. The result provides real-model evidence that computation depth should not necessarily be treated as a fixed architectural constant.

\subsubsection{Controlled Progressive-Depth Study}

To isolate the underlying phenomenon, we construct a controlled progressive-depth setting. The objective is to test whether variable optimal depth can arise even when the data-generating process is known and controlled, rather than being an artifact of a particular real benchmark.

Each sample $x_i$ has a binary label $y_i \in \{0,1\}$ and a hidden required depth $L_i^*$. The required depth represents the first layer at which the sample contains enough evidence for a reliable prediction. For a model with $L$ layers, each intermediate layer $\ell \in \{1,\ldots,L\}$ produces a prediction probability
\begin{equation}
    p_{i,\ell} = f_{\ell}(x_i),
\end{equation}
where $f_{\ell}$ denotes the model's prediction head applied at depth $\ell$.

The synthetic evidence is organized in three phases:
\begin{itemize}[leftmargin=*]
    \item Before the required depth, $\ell < L_i^*$, the sample provides only weak evidence, so early predictions are uncertain.
    \item At the required depth, $\ell = L_i^*$, the sample provides a strong identifying signal for the label.
    \item After the required depth, $\ell > L_i^*$, additional layers receive zero-mean refinement noise, so deeper computation is not guaranteed to improve the prediction.
\end{itemize}

For each intermediate prediction, we compute binary cross-entropy
\begin{equation}
    \operatorname{BCE}(y_i, p_{i,\ell})
    =
    -y_i\log p_{i,\ell} - (1-y_i)\log(1-p_{i,\ell}).
\end{equation}
An oracle chooses the best exit depth for each sample by minimizing this loss:
\begin{equation}
    \hat{d}_i
    =
    \arg\min_{\ell \in \{1,\ldots,L\}}
    \operatorname{BCE}(y_i, p_{i,\ell}).
\end{equation}
Here, $\hat{d}_i$ is the empirically best depth for sample $i$, which may differ from the hidden required depth $L_i^*$ because the model's intermediate predictions can be affected by noise and imperfect learning.

We compare this oracle variable-depth policy with fixed full-depth computation. Let $N$ be the number of samples. The full-depth loss is
\begin{equation}
    \mathcal{L}_{\mathrm{full}}
    =
    \frac{1}{N}\sum_{i=1}^{N}
    \operatorname{BCE}(y_i, p_{i,L}),
\end{equation}
and the oracle variable-depth loss is
\begin{equation}
    \mathcal{L}_{\mathrm{oracle}}
    =
    \frac{1}{N}\sum_{i=1}^{N}
    \operatorname{BCE}(y_i, p_{i,\hat{d}_i}).
\end{equation}
We report oracle gain as
\begin{equation}
    \operatorname{Gain}
    =
    \mathcal{L}_{\mathrm{full}} - \mathcal{L}_{\mathrm{oracle}}.
\end{equation}
A positive gain means that choosing the best depth per sample improves prediction quality relative to always using the final layer.

We evaluate 4-, 8-, and 12-layer models; three difficulty distributions; 32,768 samples per condition; and nine total model/difficulty conditions. The uniform difficulty distribution spreads required depths across the full model, the early-heavy distribution places more mass on shallower required depths, and the late-heavy distribution places more mass on deeper required depths.

In Table~\ref{tab:progressive-depth}, ``Unique best depths'' counts how many distinct values appear among the oracle-selected depths $\{\hat{d}_i\}_{i=1}^{N}$. ``Best-depth mean $\pm$ std'' reports the mean and standard deviation of the oracle-selected depth distribution. ``Oracle gain'' is the BCE improvement of the oracle variable-depth policy over fixed full-depth computation.

\begin{table}[h]
\centering
\caption{Controlled progressive-depth results. Oracle gain is measured relative to fixed full-depth computation in BCE.}
\label{tab:progressive-depth}
\small
\begin{tabular}{llccc}
\toprule
Model & Difficulty & Unique best depths & Best-depth mean $\pm$ std & Oracle gain \\
\midrule
4L & Uniform & 4 & $3.24 \pm 0.98$ & 0.055 \\
4L & Early-heavy & 4 & $2.97 \pm 1.12$ & 0.054 \\
4L & Late-heavy & 4 & $3.53 \pm 0.82$ & 0.045 \\
8L & Uniform & 8 & $6.23 \pm 2.00$ & 0.106 \\
8L & Early-heavy & 8 & $5.64 \pm 2.34$ & 0.102 \\
8L & Late-heavy & 8 & $6.89 \pm 1.70$ & 0.088 \\
12L & Uniform & 12 & $9.24 \pm 2.99$ & 0.143 \\
12L & Early-heavy & 12 & $8.33 \pm 3.54$ & 0.141 \\
12L & Late-heavy & 12 & $10.23 \pm 2.58$ & 0.121 \\
\bottomrule
\end{tabular}
\end{table}

Every condition contains multiple distinct optimal depths. For the 12-layer uniform-difficulty model, all 12 candidate depths are selected by at least some samples and $\mathrm{Std}(L^*)=2.99$. The oracle variable-depth policy consistently improves over fixed full-depth computation, with gains ranging from 0.045 to 0.143 BCE.

Across model depths and difficulty regimes, different samples have different optimal computation depths, and adapting depth can improve both computation efficiency and prediction quality. This motivates the second design principle of \method: computation depth should be determined by the maturity of each input rather than fixed globally.

\section{Additional Ablation and Policy Analyses}

\subsection{Remaining Design Choices}
\label{sec:ablation-design}

\paragraph{Discrete versus graded specialization.}
\Cref{eq:gate_prob} switches discretely between shared and specialized computation rather than interpolating between them.
\Cref{tab:ablation-design} compares this discrete gate with a graded alternative that mixes the shared and specialized outputs by a continuous weight, and also ablates the action-mixture target inside $\Ls_{\mathrm{gate}}$, the expert count $E$, and the routing width $k$.
Removing the action-mixture target tests the failure mode it is meant to prevent: gate saturation, where the shared expert is abandoned and the model degenerates into a fully specialized backbone that has lost cross-functional transfer.

\paragraph{Expert count and routing width.}
Widening the pool to $E=6$ experts edges out the deployed $E=4,k=2$ configuration on every benchmark in \Cref{tab:ablation-design}, but \method is designed to be a lightweight adapter, and $E=6$ adds proportionally more trained parameters than $E=4$'s $0.381\%$ overhead (\Cref{tab:hparams}) for a marginal accuracy gain. We keep $E=4$ as the deployed setting and read the $E=6$ result as headroom rather than a reason to grow the added parameter budget. Shrinking the pool to $E=2,k=2$ instead removes the shared-versus-specialized choice at each routed layer, which costs more than widening does, confirming that a nontrivial expert pool is needed for the router to specialize meaningfully.

\begin{table}[h]
\centering
\caption{Capacity-axis design ablations on Bagel. ``Rel. compute'' is the normalized route/action proxy defined in \Cref{sec:metrics}.}
\label{tab:ablation-design}
\small
\setlength{\tabcolsep}{4pt}
\resizebox{\linewidth}{!}{
\begin{tabular}{lcccccc}
\toprule
Variant & Rel. compute & MMMU & MMBench-CN & MMBench-EN & MathVista & MMStar \\
\midrule
Graded mixing instead of discrete gate  & 0.987 & 54.8 & 82.4 & 85.2 & 70.2 & 69.6 \\
No action-mixture target in $\Ls_{\mathrm{gate}}$ & 0.988 & 55.2 & 82.0 & 84.0 & 72.2 & 70.4 \\
$E=4$, $k=1$ & 0.958 & 54.0 & 85.87 & 86.75 & 73.4 & 70.93 \\
$E=2,k=2$ & 0.978 & 54.4 & 82.0 & 84.8 & 72.4 & 70.2 \\
$E=6,k=2$  & 0.951 & 55.2 & 86.6 & 87.1 & 73.6 & 74.4 \\
\bottomrule
\end{tabular}}
\end{table}

\clearpage
\subsection{Early-Exit Subset Audit}
\label{sec:appendix-understanding}

The main paper already reports the overall selected-depth distribution, so we
do not repeat that plot here.  Instead, we analyze the rare early-exit cases
in the same four-GPU lightweight screen.  The screen produced 16 raw routes
with a selected depth below 28.  Four-way circular choice evaluation creates
duplicate MMBench routes, leaving 10 unique inputs after deduplication: three
MMMU inputs, five MathVista inputs, and two MMBench-English inputs.

Every unique early case selected depth 26; no case selected depth 22 or 24.
This is a targeted audit of the early-exit tail, not a representative estimate
of the full benchmark distribution.  The evaluators do not provide a common
per-example correctness field across all three suites, so the analysis focuses
on the input and router signals rather than claiming early-exit accuracy.

\subsubsection{What receives an early exit?}

Table~\ref{tab:early-exit-cases} lists every unique early case.  The subset is
concentrated in a few visually grounded or short-answer task families:
MathVista icon-style questions account for four of the five MathVista cases,
while the MMMU cases come from biology and chemistry.  Question length is not
uniform: the subset contains very short 21-character questions as well as a
726-character biology question.  Thus the observed early exits cannot be
explained by a simple short-question rule.

\begin{table}[H]
\centering
\caption{Unique early-exit cases in the lightweight screen. MMBench circular-choice copies are deduplicated. The router selected depth 26 in every case; the signals are recorded at the selected depth.}
\label{tab:early-exit-cases}
\scriptsize
\setlength{\tabcolsep}{3pt}
\begin{tabular}{lllrrrrr}
\toprule
Suite & Case & Task & Chars. & Depth & Halt@26 & Risk@26 & Regret@26 \\
\midrule
MMBench-EN & 493 & MMBench-English & 66 & 26 & 0.581 & 0.091 & 0.038 \\
MMBench-EN & 630 & MMBench-English & 41 & 26 & 0.640 & 0.081 & 0.030 \\
MMMU & Biology\_21 & biology & 218 & 26 & 0.511 & 0.098 & 0.045 \\
MMMU & Biology\_8 & biology & 726 & 26 & 0.506 & 0.088 & 0.035 \\
MMMU & Chemistry\_30 & chemistry & 257 & 26 & 0.505 & 0.090 & 0.037 \\
MathVista & 129 & iconqa\_integer & 104 & 26 & 0.523 & 0.087 & 0.034 \\
MathVista & 157 & iconqa\_integer & 21 & 26 & 0.524 & 0.097 & 0.044 \\
MathVista & 163 & iconqa\_integer & 21 & 26 & 0.535 & 0.096 & 0.043 \\
MathVista & 187 & iconqa\_integer & 21 & 26 & 0.535 & 0.097 & 0.043 \\
MathVista & 92 & vqa2\_0\_text & 60 & 26 & 0.602 & 0.093 & 0.040 \\
\bottomrule
\end{tabular}
\end{table}

\begin{figure}[H]
  \centering
  \includegraphics[width=.94\linewidth]{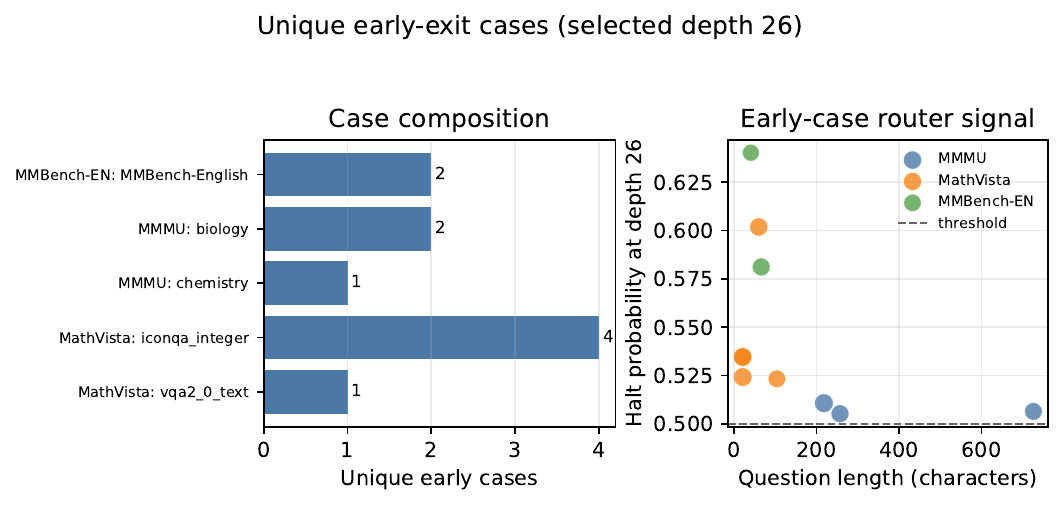}
  \caption{Early-exit subset composition and router signal.  Each point is a
  unique input selected at depth 26; marker size is proportional to the
  residual-risk upper bound.}
  \label{fig:early-exit-profile}
\end{figure}

Figure~\ref{fig:early-exit-profile} summarizes the subset composition and
plots question length against the halt probability at the selected depth.
All points lie at or above the router's 0.5 threshold, while their marker size
encodes the residual-risk upper bound.  The small sample and repeated task
families make this an audit of behavior, not a statistical generalization.

\subsubsection{Router traces around the early decision}

\begin{figure}[H]
  \centering
  \includegraphics[width=.90\linewidth]{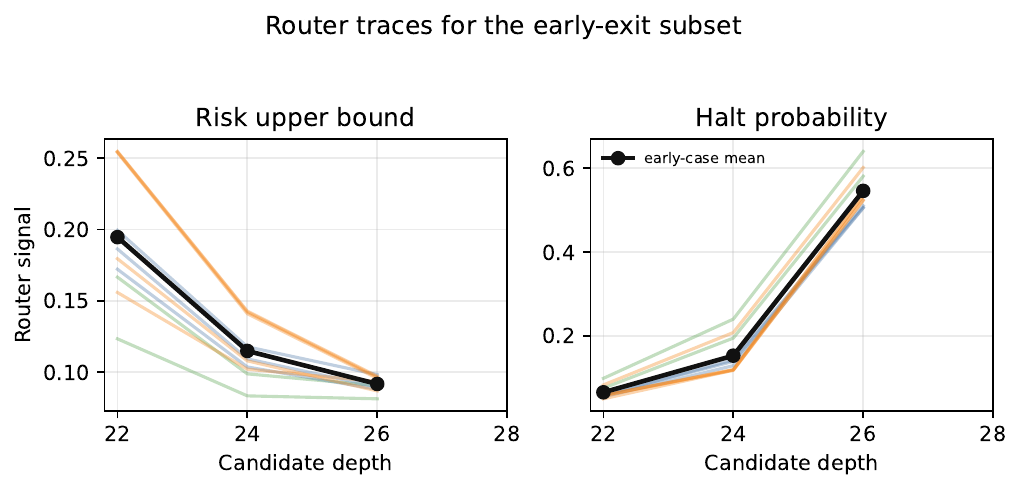}
  \caption{Candidate-depth router traces for the unique early-exit subset.
  Faint lines are individual cases and the black line is their mean at each
  evaluated candidate depth.}
  \label{fig:early-exit-traces}
\end{figure}

Figure~\ref{fig:early-exit-traces} follows the complete router trace for each
unique early case.  The individual traces show a consistent pattern: the
risk upper bound falls from depths 22 to 26, while halt probability rises and
crosses the 0.5 decision threshold at depth 26.  The router therefore exits
when its own risk signal has decreased sufficiently, rather than because the
input was assigned a fixed shallow depth in advance.  Since these traces stop
once the policy exits, there is no depth-28 point for an early case.

\subsubsection{Why the gate passes for early cases}

\begin{figure}[H]
  \centering
  \includegraphics[width=.90\linewidth]{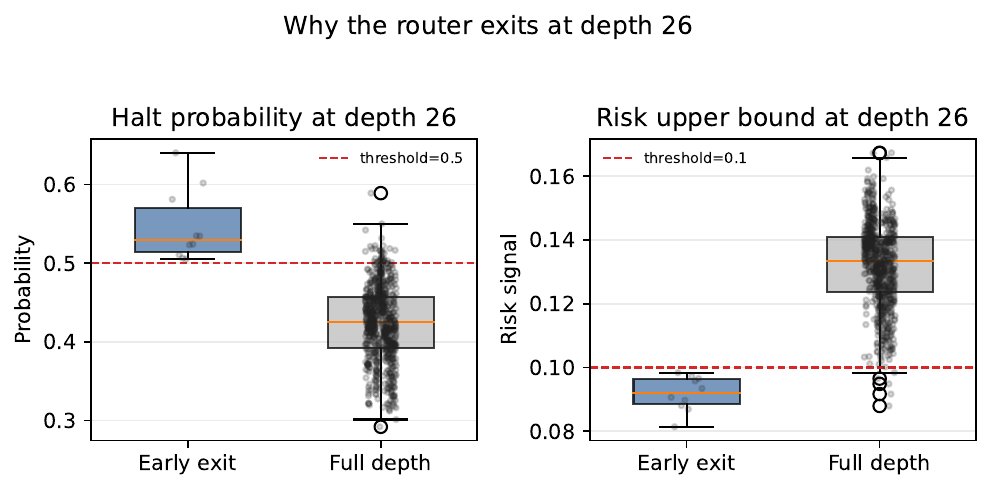}
  \caption{Depth-26 gate signals for early-exit and full-depth inputs.  The
  dashed lines show the configured halt-probability and risk thresholds.}
  \label{fig:early-exit-gate}
\end{figure}

The implementation requires both a halt probability of at least $0.5$ and a
risk upper bound of at most $0.1$ before exiting.  We compare these two gate
signals at depth 26 for every deduplicated input in the screen.  As shown in
Figure~\ref{fig:early-exit-gate} and Table~\ref{tab:early-exit-gate}, all 10
early cases satisfy both conditions, whereas none of the 740 full-depth cases
does.  Only 5.0\% of full-depth cases cross the halt threshold alone, and
their risk signal remains too high.  This provides a more mechanistic
explanation for the rare early exits than question length or task identity by
itself.

\begin{table}[H]
\centering
\caption{Depth-26 gate-condition comparison. ``Halt pass'' uses the configured halt threshold 0.5; ``Joint pass'' requires both halt probability $\geq 0.5$ and risk upper bound $\leq 0.1$.}
\label{tab:early-exit-gate}
\scriptsize
\begin{tabular}{lrrrrr}
\toprule
Group & $N$ & Mean halt@26 & Mean risk@26 & Halt pass & Joint pass \\
\midrule
early & 10 & 0.546 & 0.092 & 100.0\% & 100.0\% \\
full & 740 & 0.424 & 0.132 & 5.0\% & 0.0\% \\
\bottomrule
\end{tabular}
\end{table}

\paragraph{Interpretation.}
The early-exit subset provides concrete evidence that the learned policy is
active on individual understanding inputs, but it also clarifies the current
checkpoint's operating point: early exits are rare and all occur at depth 26.
The result supports input-conditioned routing as a mechanism while motivating
future calibration or threshold studies if broader compute savings are desired.

\section{Extended Case Study}
\label{sec:extended-case-study}

This appendix gives the full selected qualitative set referenced in
Section~\ref{sec:case-study}.  The examples are strict paired wins: the
FlexUMM answer or image is correct under the benchmark evaluator and the
baseline comparator is not.  They are selected for auditability and diversity,
and are not intended to replace aggregate benchmark results.

\subsection{Understanding cases}

Table~\ref{tab:extended-understanding} lists all ten paired understanding
examples.  The MMBench entries are representative displayed rows from the
four-way circular-group evaluation used in the paired output analysis.

\begingroup
\scriptsize
\setlength{\tabcolsep}{3pt}
\begin{longtable}{@{}c p{.22\textwidth} c c c@{}}
  \caption{Complete selected understanding set.  Each row includes the
  benchmark input image and the paired answers.  The FlexUMM answer agrees
  with the ground truth and the baseline answer does not.}
  \label{tab:extended-understanding} \\
  \toprule
  \# & Input image & FlexUMM & Baseline & GT \\
  \midrule
  \endfirsthead
  \toprule
  \# & Input image & FlexUMM & Baseline & GT \\
  \midrule
  \endhead
  \bottomrule
  \endfoot
  1 & \includegraphics[width=.14\textwidth]{figures/case_study/understanding_01_mmmu_music.png}
    & A & B & A \\
  2 & \includegraphics[width=.14\textwidth]{figures/case_study/understanding_02_mmmu_clinical.png}
    & D & B & D \\
  3 & \includegraphics[width=.14\textwidth]{figures/case_study/understanding_03_mathvista_28.png}
    & 8.5 & 7.5 & 8.5 \\
  4 & \includegraphics[width=.14\textwidth]{figures/case_study/understanding_04_mathvista_270.png}
    & $100^\circ$ & $130^\circ$ & $100^\circ$ \\
  5 & \includegraphics[width=.14\textwidth]{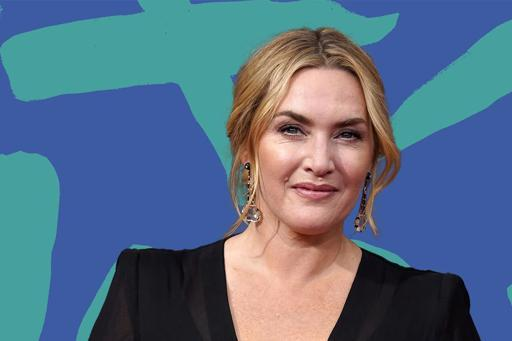}
    & C & B & C \\
  6 & \includegraphics[width=.14\textwidth]{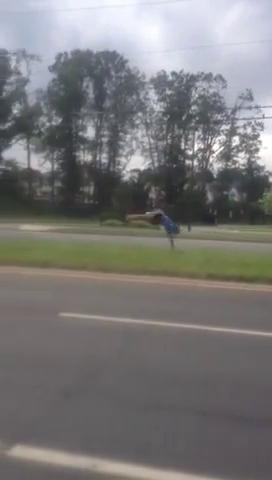}
    & A & B & A \\
  7 & \includegraphics[width=.14\textwidth]{figures/case_study/understanding_07_mmbench_cn_350.png}
    & B & A & B \\
  8 & \includegraphics[width=.14\textwidth]{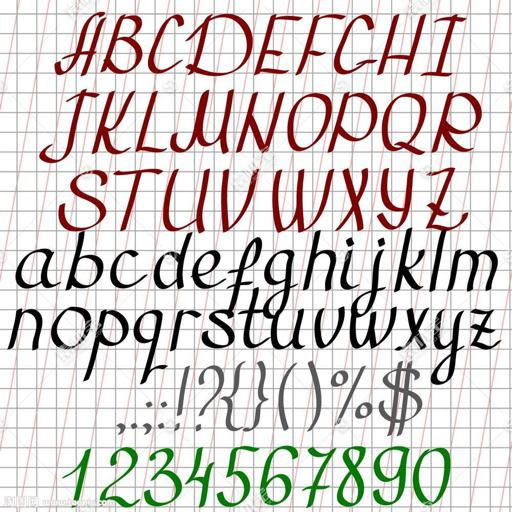}
    & B & C & B \\
  9 & \includegraphics[width=.14\textwidth]{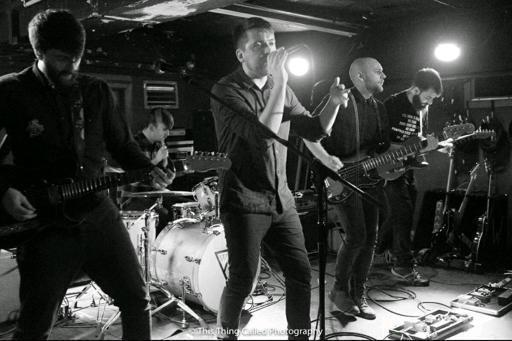}
    & D & C & D \\
  10 & \includegraphics[width=.14\textwidth]{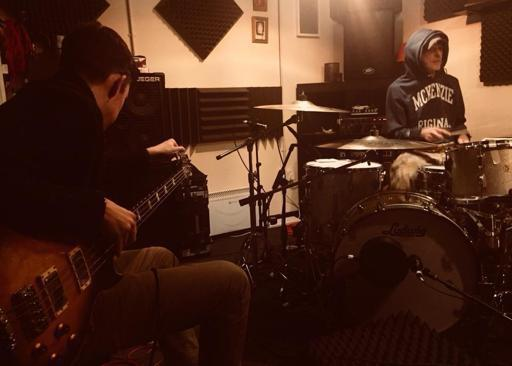}
    & B & C & B \\
\end{longtable}
\endgroup

\subsection{Generation cases}

\begin{figure}[t]
  \centering
  \includegraphics[width=\textwidth]{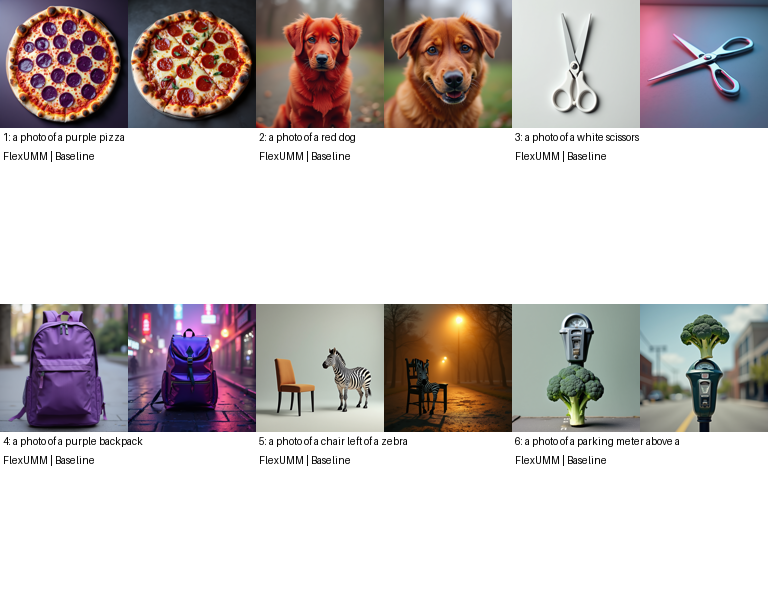}
  \caption{Selected generation set.  Each tile contains the FlexUMM
  image followed by the paired baseline image.}
  \label{fig:extended-generation}
\end{figure}

Figure~\ref{fig:extended-generation} shows six distinct selected generation
wins spanning color, object identity, and spatial-relation constraints.  The
set includes one sample for each of the repeated red-dog and white-scissors
prompts, while omitting the purple-bear samples.
Within each tile, the FlexUMM image is on the left and the baseline comparator
is on the right.  The pairs were generated with identical prompts and
evaluation settings, and the detector outcome is summarized in
Table~\ref{tab:extended-generation}.

\begin{table}[t]
  \centering
  \caption{Detector outcomes for the selected generation set.  The rows are
  individual image-level wins from distinct prompts.}
  \label{tab:extended-generation}
  \scriptsize
  \setlength{\tabcolsep}{3pt}
  \begin{tabularx}{\textwidth}{@{}c X X X@{}}
    \toprule
    \# & Prompt & FlexUMM result & Baseline failure mode \\
    \midrule
    1 & purple pizza & Pass & No purple pizza; red pizza detected \\
    2 & red dog & Pass & Brown dog rather than red \\
    3 & white scissors & Pass & Purple scissors rather than white \\
    4 & purple backpack & Pass & Suitcase/handbag detected, not backpack \\
    5 & chair left of a zebra & Pass & Bench detected; no chair \\
    6 & parking meter above a broccoli & Pass & Parking meter below the broccoli \\
    \bottomrule
  \end{tabularx}
\end{table}

\section{Limitations}
\label{sec:limitations}

Our evaluation considers a finite set of candidate depths for each backbone, which limits the granularity of the available quality-compute operating points. In addition, our experiments focus on a limited set of unified multimodal backbones and model scales.
Extending the evaluation to finer-grained depth configurations and a broader range of architectures could further characterize the flexibility and applicability of the proposed framework.

\end{document}